\documentclass{article}

\usepackage[preprint]{neurips_2025}

\usepackage[utf8]{inputenc} 
\usepackage[T1]{fontenc}    
\usepackage{hyperref}       
\usepackage{url}            
\usepackage{booktabs}       
\usepackage{amsfonts}       
\usepackage{nicefrac}       
\usepackage{microtype}      
\usepackage{xcolor}         

\usepackage{graphicx}
\usepackage{acronym}
\usepackage{amsfonts}
\usepackage{enumitem}
\usepackage{subcaption}
\usepackage{multirow}
\usepackage{algorithm}
\usepackage{algpseudocode}
\usepackage{amsmath}

\acrodef{TSCH}{Time-Slotted Channel Hopping}
\acrodef{RF}{Radio Frequency}
\acrodefindefinite{RF}{an}{a}
\acrodef{LNA}{Low-Noise Amplifier}
\acrodefindefinite{LNA}{an}{a}
\acrodef{LO}{Local Oscillator}
\acrodefindefinite{LO}{an}{a}
\acrodef{ADC}{Analog-to-Digital Converter}
\acrodef{IF}{Intermediate Frequency}
\acrodef{CMOS}{Complementary Metal-Oxide-Semiconductor}
\acrodef{DBP}{Digital Baseband Processor}
\acrodef{DSSS}{Direct Sequence Spread Spectrum}
\acrodef{ASK}{Amplitude Shift Keying}
\acrodef{MSK}{Minimum Shift Keying}
\acrodefindefinite{MSK}{an}{a}
\acrodef{FSK}{Frequency Shift Keying}
\acrodefindefinite{FSK}{an}{a}
\acrodef{oqpsk}[O-QPSK]{Offset-Quadrature Phase Shift Keying}
\acrodef{BPF}{Band-Pass Filter}
\acrodef{PRR}{Packet Reception Ratio}
\acrodef{SDR}{Software Defined Radio}
\acrodefindefinite{SDR}{an}{a}
\acrodef{IC}{Integrated Circuit}
\acrodef{WSN}{Wireless Sensor Networks}
\acrodef{CCA}{Clear Channel Assessment}
\acrodef{MAC}{Medium Access Control}
\acrodef{IoT}{Internet of Things}
\acrodef{COTS}{Commercial Off-The-Shelf}

\acrodef{CO}{Constraint Optimizer}
\acrodef{COP}{Combinatorial Optimization Problem}
\acrodef{EVD}{Eigenvalue Decomposition}
\acrodef{ML}{Machine Learning}
\acrodef{MLP}{Multilayer Perceptron}
\acrodef{PE}{Positional Encoding}
\acrodefindefinite{ML}{an}{a}
\acrodef{DL}{Deep Learning}
\acrodef{GRL}{Graph Representation Learning}
\acrodef{GNN}{Graph Neural Network}
\acrodef{BN}{Bayesian Network}
\acrodef{RNN}{Recurrent Neural Network}
\acrodefindefinite{RNN}{an}{a}
\acrodef{NMT}{Neural Machine Translation}
\acrodefindefinite{NMT}{an}{a}
\acrodef{RL}{Reinforcement Learning}
\acrodefindefinite{RL}{an}{a}
\acrodef{DRL}{Deep Reinforcement Learning}
\acrodef{SSL}{Self-Supervised Learning}
\acrodefindefinite{SSL}{an}{a}
\acrodef{seq2seq}{Sequence-to-Sequence}
\acrodef{LSTM}{Long Short-Term Memory}
\acrodefindefinite{LSTM}{an}{a}
\acrodef{KDE}{Kernel Density Estimator}
\acrodef{LLM}{Large Language Model}
\acrodef{CDF}{Cumulative Distribution Function}

\title{CASTILLO: Characterizing Response Length Distributions of Large Language Models}

\author{%
  Daniel F. Perez-Ramirez \\
  RISE Computer Science \\ KTH Royal Institute of Technology\\
  Stockholm, Sweden \\
  \texttt{daniel.perez@ri.se -- dfpr@kth.se} \\
  \And
  Dejan Kostic \\
  KTH Royal Institute of Technology \\ RISE Computer Science \\
  Stockholm, Sweden \\
  \texttt{dmk@kth.se} \\
  \And
  Magnus Boman \\
  Karolinska Institutet \\
  Solna, Sweden \\
  \texttt{magnus.boman@ki.se} \\
}

\begin{document}

\maketitle

\begin{abstract}
    Efficiently managing compute resources for Large Language Model (LLM) inference remains challenging due to the inherently stochastic and variable lengths of autoregressive text generation. 
    Accurately estimating response lengths in advance enables proactive resource allocation, yet existing approaches either bias text generation towards certain lengths or rely on assumptions that ignore model- and prompt-specific variability. 
    We introduce CASTILLO, a dataset characterizing response length distributions across 13 widely-used open-source LLMs evaluated on seven distinct instruction-following corpora.
    For each $\langle$prompt, model$\rangle$ sample pair, we generate 10 independent completions using fixed decoding hyper-parameters, record the token length of each response, and publish summary statistics (mean, std-dev, percentiles), along with the shortest and longest completions, and the exact generation settings.
    Our analysis reveals significant inter- and intra-model variability in response lengths (even under identical generation settings), as well as model-specific behaviors and occurrences of partial text degeneration in only subsets of responses.
    CASTILLO enables the development of predictive models for proactive scheduling and provides a systematic framework for analyzing model-specific generation behaviors. 
    We publicly release the dataset and code to foster research at the intersection of generative language modeling and systems.
    
\end{abstract}

    


\section{Introduction}

\acp{LLM} have  transformed natural language processing by enabling general-purpose language understanding and generation at unprecedented scale~\citep{brown2020llmsfewshot, raffel2020exploring, chowdhery2023palm}. 
Instruction-tuned LLMs are now widely deployed across a range of applications—including question answering, dialogue systems, and code generation—powering production services used by millions of users daily~\citep{ouyang2022instructgpt, chen2021codex, achiam2023gpt4, team2023gemini}. However, their computational and memory demands pose significant challenges to scalable and cost-effective inference in production systems.

A central bottleneck in LLM serving systems is the difficulty of managing compute and memory resources efficiently, especially under high concurrency and latency constraints. 
Recent systems research has emphasized memory management of the attention Key-Value (KV) cache~\citep{kwon2023vllm} and reactive scheduling strategies that respond to runtime demand fluctuations~\citep{duan2024muxserve, patel2024splitwise, agrawal2024sarathi}. 
While effective to some degree, reactive methods are inherently limited by their inability to anticipate the variable and stochastic nature of autoregressive generation in transformer-based models~\citep{vaswani2017attention, holtzman2020textdegeneration}. This unpredictability leads to inefficient resource utilization and increased operational overhead.

\begin{figure}
    \centering
    \includegraphics[width=0.98\linewidth]{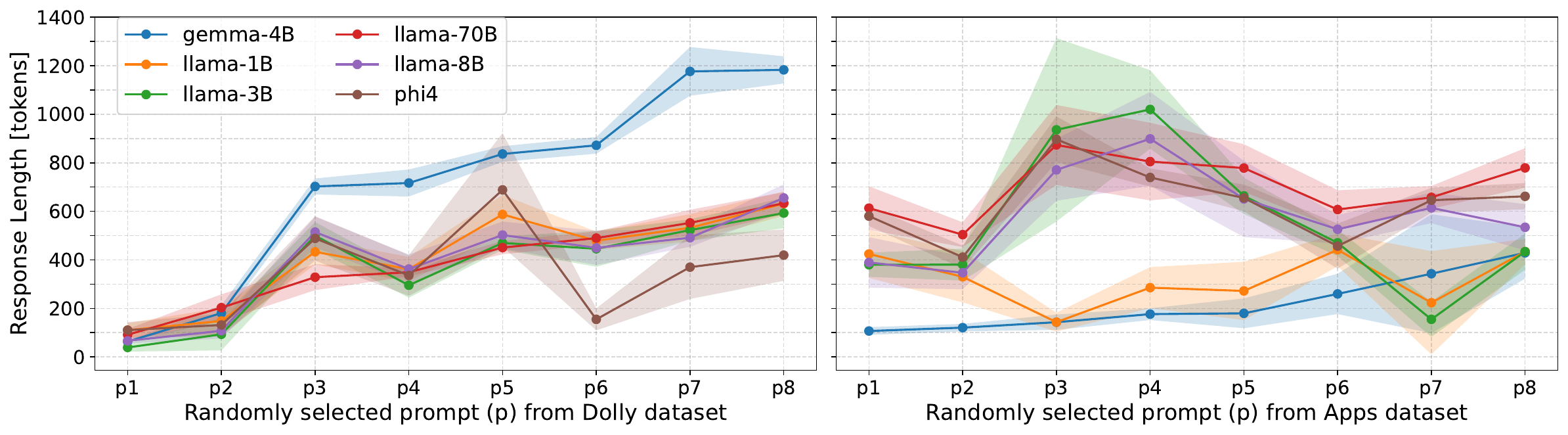}
    \caption{\textbf{LLM response lengths exhibit significant inter- and intra-model variability}. Points depict the mean response length for a ⟨prompt, model⟩ pair; shaded regions depict ±1 standard deviation over 10 independent responses under fixed decoding parameters. Significant variation both among different models and within the 10 sampled responses, even within the same model.}
    \label{fig:intro_plot}
\end{figure}

Proactive scheduling strategies, with the aim to predict response lengths before generation, have the potential to significantly improve \acp{LLM} inference efficiency. Existing approaches often rely on strong assumptions, such as fixed-length outputs across prompts or models~\citep{jin2023s3, shahout2025trail}, or introduce unwanted bias by implicitly conditioning the generation toward target lengths~\citep{zheng2023response}. 
Figure~\ref{fig:intro_plot} highlights the substantial variability in response lengths across both prompts and models. Even under fixed decoding parameters, we observe that no model consistently produces longer or shorter outputs—one model may respond concisely to a particular prompt but verbosely to another. Even when fixing the model and prompt, we see cases of high variance among responses that can extend for hundreds of tokens for some models (see Apps-p3 in Figure~\ref{fig:intro_plot}). This unpredictability underscores the limitations of using only the text prompt alone to estimate response lengths.

In this work, we present \textbf{CASTILLO}--\textbf{C}har\textbf{A}cterizing re\textbf{S}ponse leng\textbf{T}h d\textbf{I}stributions in \textbf{L}arge \textbf{L}anguage m\textbf{O}dels--a large-scale dataset designed to empirically characterize LLM response length distributions across diverse models and tasks.
CASTILLO spans 13 widely-used open-source \acp{LLM} and 7 instruction-following datasets, covering a diverse range of model architectures, sizes, and text corpora. For each ⟨prompt, model⟩ sample pair, we generate 10 independent completions using fixed decoding hyper-parameters and record the token length of each response. We release summary statistics (mean, standard deviation, percentiles), along with the shortest and longest completions and the exact generation settings.

Our analysis reveals several key insights:
\begin{enumerate}[leftmargin=2em]
    \item \textbf{Inter-model variability}: Different models exhibit markedly different response length distributions, even when prompted with identical inputs (see Figures~\ref{fig:heatmaps} and ~\ref{fig:box-bar_model_dataset}). This underscores the need for model-specific characterization of generation behavior.
    \item \textbf{Intra-model variability}: within a single model, we observe high variance in response lengths across different prompts (see Figure~\ref{fig:coeff-var}), even with fixed decoding settings, highlighting the influence of prompt semantics on generation length.
    \item \textbf{Partial text degeneration}: In some cases, a subset of the 10 completions for a given prompt-model pair exhibit text degeneration artifacts (e.g., repetition or incoherence), suggesting localized instability not captured by single-sample evaluations (see Figure~\ref{fig:textdegen}).
\end{enumerate}

CASTILLO serves as both a benchmarking resource and a foundation for building predictive models of response length. Such models could inform proactive scheduling strategies, reduce latency, and optimize resource allocation in LLM serving infrastructures. Additionally, CASTILLO provides a controlled framework for systematically analyzing and comparing generation behaviors across different models, prompts, and decoding settings.
We release the dataset and codebase (see Appendix~\ref{app:availability}) to support reproducibility and foster future work at the intersection of machine learning and systems.

\section{Background and Related Work}

To our knowledge, CASTILLO is the first publicly-available dataset to empirically characterize response length distributions across multiple LLMs and instruction-following datasets.

\begin{figure}
    \centering
    \includegraphics[width=0.98\linewidth]{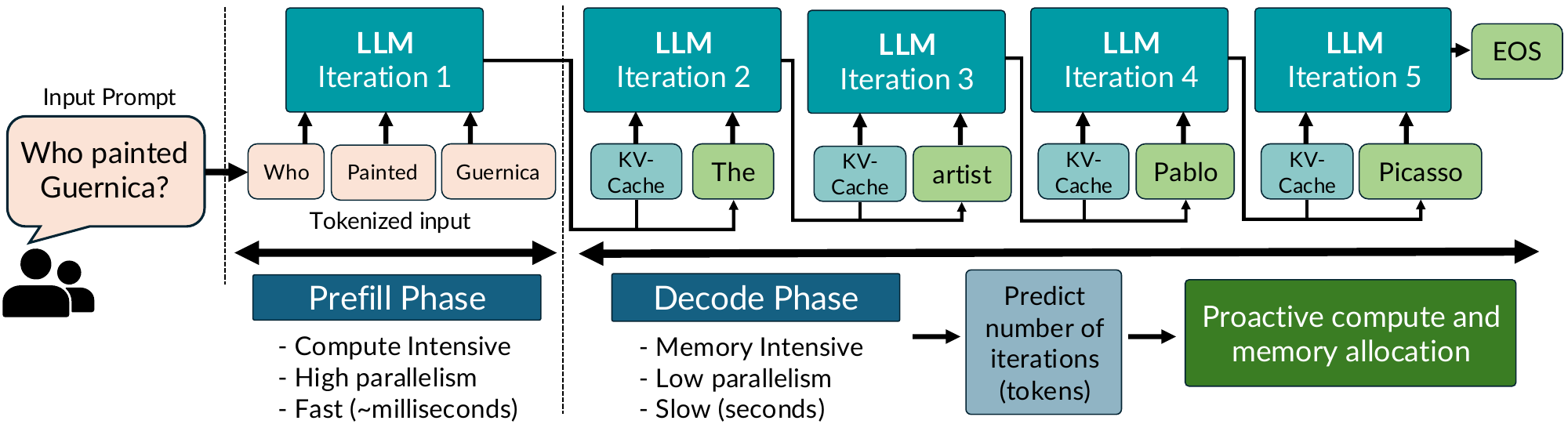}
    \caption{Transformer-based LLM inference in production is divided into a fast and compute intensive phase (prefill) and a time-consuming and memory intensive phase (decode).}
    \label{fig:prefill-decode}
\end{figure}

\paragraph{Transformer-based models} Modern \acp{LLM} are predominantly based on the Transformer architecture~\citep{vaswani2017attention}, which models text as sequences of discrete tokens and generates outputs through an autoregressive process--predicting each token conditioned on the preceding context. This token-by-token generation is inherently stochastic and influenced by both the model parameters and sampling hyperparameters (such as temperature) and decoding configuration~\citep{holtzman2020textdegeneration}. In this work, we focus exclusively on transformer-based \acp{LLM}, given their ubiquity and dominance in both academic and industrial deployments.

\paragraph{Efficient \ac{LLM} inference} In production systems, LLM inference can be decomposed into two distinct phases: a prefill phase and a decoding phase~\citep{agrawal2023sarathi, patel2024splitwise, agrawal2024sarathi}. Figure ~\ref{fig:prefill-decode} depicts the process: the prefill phase encodes the input prompt and computes the KV-Cache for the attention layers--a compute-intensive parallelizable operation that typically executes in the millisecond range. The decoding phase then generates tokens autoregressively, both using and augmenting the KV-Cache, incurring sequential dependencies and high memory overhead. This phase often dominates latency, stretching into seconds or even minutes for long responses. Meanwhile, scheduling decisions--such as request admission or GPU placement--typically occur in the microsecond range, creating a critical need for accurate, low-latency predictions of output length to enable proactive scheduling.

Recent work aims at improving \ac{LLM} inference efficiency through systems-level innovations. Methods such as PagedAttention~\citep{kwon2023vllm} and FlashAttention~\citep{dao2022flashattention} address memory locality and computation speed during attention computation. Inference engines like vLLM~\citep{kwon2023vllm}, SplitWise~\citep{patel2024splitwise}, Sarathi~\citep{agrawal2024sarathi}, Flexgen~\citep{sheng2023flexgen}, and MuxServe~\citep{duan2024muxserve} implement dynamic batching, cache eviction policies, and token-level scheduling. CASTILLO is orthogonal to these efforts: our work provides a standardized dataset to inform and augment such systems via better response length estimation.

\paragraph{Response Length Prediction} Recent works attempt to directly predict response lengths in advance. \cite{zheng2023response} propose to fine-tune \acp{LLM} with an auxiliary task that predicts the number of output words before generating a response. However, this method introduces length biases and was evaluated on only a single dataset, limiting generality. \cite{jin2023s3} introduce $S^3$, a classification-based predictor using DistilBERT to assign each input prompt to one of 10 output-length buckets. While efficient, $S^3$ ignores intra-model variability (each ⟨prompt, model⟩ pair has a single estimated length) and assumes inter-model homogeneity by using the same predictor for all \acp{LLM}. Moreover, the use of DistilBERT constrains prompt length to 512 tokens, and the fixed bucket resolution may be too coarse for fine-grained scheduling.
Building on $S^3$, \cite{shahout2025trail} propose TRAIL, which refines predictions during decoding by conditioning on \textit{specific} intermediate activations from the LLM and updating estimates via a Bayesian model. While effective, TRAIL introduces communication overhead and requires in-depth layer-wise model-specific analysis, which limits its generalization. 

CASTILLO contributes to this growing area by offering a large-scale dataset that captures both inter- and intra-model response variability under fixed decoding conditions. By releasing response statistics (mean, percentiles, extremes), generation hyperparameters, and evidence of partial degeneration, our dataset enables systematic study and benchmarking of predictive length estimation methods, and serves as a robust testbed for building more general and proactive inference schedulers.

\section{CASTILLO Dataset Construction}
\label{sec:dataset}

This section details the construction of CASTILLO, including source dataset selection, open-source \acp{LLM} used, and the response generation framework. Our dataset is designed to enable empirical study of response length variability and support the development of predictor models for proactive inference scheduling.

\subsection{Source Dataset Selection}
\label{subsec:source-datasets}

To capture a diverse range of instruction-following behaviors, we consider seven publicly available and widely used open-source datasets spanning both general-purpose NLP tasks and code generation benchmarks. Our selection includes open-ended instruction datasets---\textbf{Dolly}, \textbf{ShareGPT}, and \textbf{Alpaca}---as well as structured problem-solving and code-oriented datasets such as \textbf{Mbpp}, \textbf{Apps}, \textbf{DS-1000}, and \textbf{BigCodeBench}. All selected datasets have seen prior adoption in peer-reviewed machine learning or systems research and are representative of real-world \ac{LLM}workloads.

Each dataset is processed independently. When a dataset contains more than 2000 samples, we randomly downsample to a uniform cap of 2000 prompts. For datasets with internal categorization (e.g., by task type or domain), we apply stratified sampling to maintain representative distributions. Importantly, we verify that the downsampled subsets preserve the original distributions of input token lengths, as shown in Appendix~\ref{app:sourcedatasets}. The resulting prompts are partitioned into training, validation, and test splits using a 70/20/10 ratio, with optional stratification where applicable. The training and validation sets are intended for the development of response length predictors, while the test set is reserved for system-level evaluation.

Table~\ref{table:raw_datasets} presents an overview of the seven datasets considered, including token-level statistics of the input prompts after tokenization using the \texttt{llama-3.2-1B} tokenizer. We report the mean, minimum, percentiles (P25, P50, P75, P99), and maximum token lengths. These statistics highlight the high diversity in prompt lengths across datasets--from compact, instruction-like inputs in Alpaca and Dolly, to longer, more verbose examples in Apps and DS-1000---underscoring the importance of considering a wide range of source datasets for building characterizing response lengths.

\subsection{LLM Model Selection}
\label{subsec:models}

To explore variability across models and architectures, CASTILLO includes 13 open-source instruction-tuned \acp{LLM} from four major organizations. 
We prioritize models that are (1) publicly available on Hugging Face, (2) used in recent systems and ML research focused on inference optimization, and (3) realistically deployable by research labs and Small- and Medium Enterpises (SMEs), particularly in the mid-size parameter range.
Our selection includes models from the LLaMA, Mistral, Qwen, Phi, and Gemma families, ranging in size from 1B to 70B parameters. By including multiple sizes within families, we allow for both intra-family and inter-family comparisons of response behavior. We query each model using its instruction-tuned variant and standardized HuggingFace model configuration. Appendix~\ref{app:llmsused} (Table~\ref{tab:llms}) lists the specific model identifiers, their parameter sizes, organizations, and the number of H100 GPUs used for inference.

\begin{table}
  \caption{Source datasets used by CASTILLO to generate input prompts for the \acp{LLM}, with sample counts and token length statistics after tokenization with Llama-3.2-1B. "P{X}" denotes the X\textsuperscript{th} percentile. More details about source dataset analysis and processing in Appendix~\ref{app:sourcedatasets}.}
  \label{table:raw_datasets}
  \centering
  \small
  \begin{tabular}{lcrrrrrrrr}
    \toprule
    Name            & Samples   & Mean   &Min. &P25 &P50  &P75 &P99  & Max. \\
    \midrule
    DollyDataset~\citep{DatabricksBlog2023DollyV2}
                    & 2000      &125.9     &36       &44     &50     &146    &795.2      &4003\\
    ShareGPT~\citep{chiang2023vicuna}    
                    & 2000      &260.5     &36       &48     &64     &168    &2534.0      &2534\\
    Alpaca~\citep{taori2023alpaca}
                    & 2000      &53.7      &39       &45     &49     &57     &114.0       &397\\
    Mbpp~\citep{austin2021mbpp}
                    & 974      &153.5      &88       &109    &131    &173    &336.3     &2265\\
    Apps~\citep{hendrycks2021apps}
                    & 2000     &545.0        &87       &307.7    &441    &650    &2105.0     &2534\\
    DS-1000~\citep{lai2023ds1000}
                    & 1000     &317.2      &67       &170.5    &283    &395    &1018.3     &2109\\
    BigCodeBench~\citep{zhuo2025bigcodebench}
                    & 1140     &179.8      &87       &137    &164    &205    &398.4     &1251\\
    \bottomrule
  \end{tabular}
\end{table}

\subsection{Response Generation Process}
\label{subsec:generation}

We implement dataset-specific prompt formatting routines to respect the schema and intended use cases of each source dataset. These routines generate consistent and model-compatible chat prompts, as detailed in the subsections of Appendix~\ref{app:sourcedatasets}.

To ensure compatibility with all models and control GPU memory usage, we cap input prompt lengths at 2500 tokens. This threshold was selected based on analysis of the input token distributions in Table~\ref{table:raw_datasets}, and strikes a balance between empirical coverage and feasibility of computing response lengths with our available resources. Notably, this limit is significantly larger than the 99th percentile token length across all datasets and avoids out-of-memory failures while preserving most of the input space relevant for inference.

For each $\langle$prompt, model$\rangle$ pair, we first transform the text prompt into a \texttt{chat\_template} using the respective model's tokenizer, we construct a batch of 10 identical prompt entries, and perform independent decoding for each using Hugging Face’s Transformers interface~\citep{wolf2020huggingface}. Generation is performed with \texttt{do\_sample=True}, and the decoding parameters (\texttt{temperature}, \texttt{top-k}, \texttt{top-p}) are recorded and stored for each sample. Our codebase (see Appendix~\ref{app:availability}) is designed to be extensible, allowing users to augment the dataset with additional models and decoding configurations by specifying them at runtime.
To guard against resource exhaustion due to pathological completions (repeated test generation in loop), we cap the maximum output length at 15000 per response (approximately 11000 words). This decision was informed by early observations of text degeneration, where some \acp{LLM} entered repetitive loops or verbose rambling that consumed the full context window and available GPU memory.

After generation, each batch of 10 completions is post-processed to extract the \emph{token length} of each response. For every $\langle$prompt, model$\rangle$ pair, we compute and store the following statistics: i) mean, standard deviation (std-dev), and percentile values (P25, P50, P75, P99) of response lengths, ii) the shortest and longest completions (cached as raw text, and iii) full decoding configuration (\texttt{temperature}, \texttt{top-k}, \texttt{top-p}). More details about the complete dataset schema can be found in Table~\ref{tab:dataset-schema} in the Appendix. 

\subsection{Sanitizing the CASTILLO Dataset}

To support downstream analysis and robustness studies, we further partition the dataset into two complementary subsets:
\begin{enumerate}
    \item A \textbf{sanitized version}, sanitizing or excluding all completions exhibiting degeneration artifacts (e.g., high repetition, loss of coherence, maximum-length responses), and
    \item A \textbf{degeneration-only subset}, containing flagged cases for which at least one of the 10 completions showed evidence of text degeneration.
\end{enumerate}

For full details and illustrative examples, see Appendix~\ref{app:textdegeneration}. 
For identifying text degeneration in CASTILLO, we applied a two-stage filtering strategy based on output length statistics. Specifically, we flagged samples as degenerated if they contained any outputs that reached the generation cap of 15,000 tokens, or if they exhibited high variance in response lengths—defined as a standard deviation greater than twice the mean combined with at least one output exceeding 8,500 tokens. This approach enabled us to distinguish true degeneration from benign verbosity and to construct both degenerated and sanitized subsets of the dataset used in our analysis. 
These two versions allow researchers to isolate text-degeneration behavior or train predictors on clean, high-quality generations.

\section{Characterizing LLM Response Distributions }
\label{sec:analysis}

\begin{figure}
    \centering
    \begin{subfigure}[b]{0.49\textwidth}
    \includegraphics[width=\linewidth]{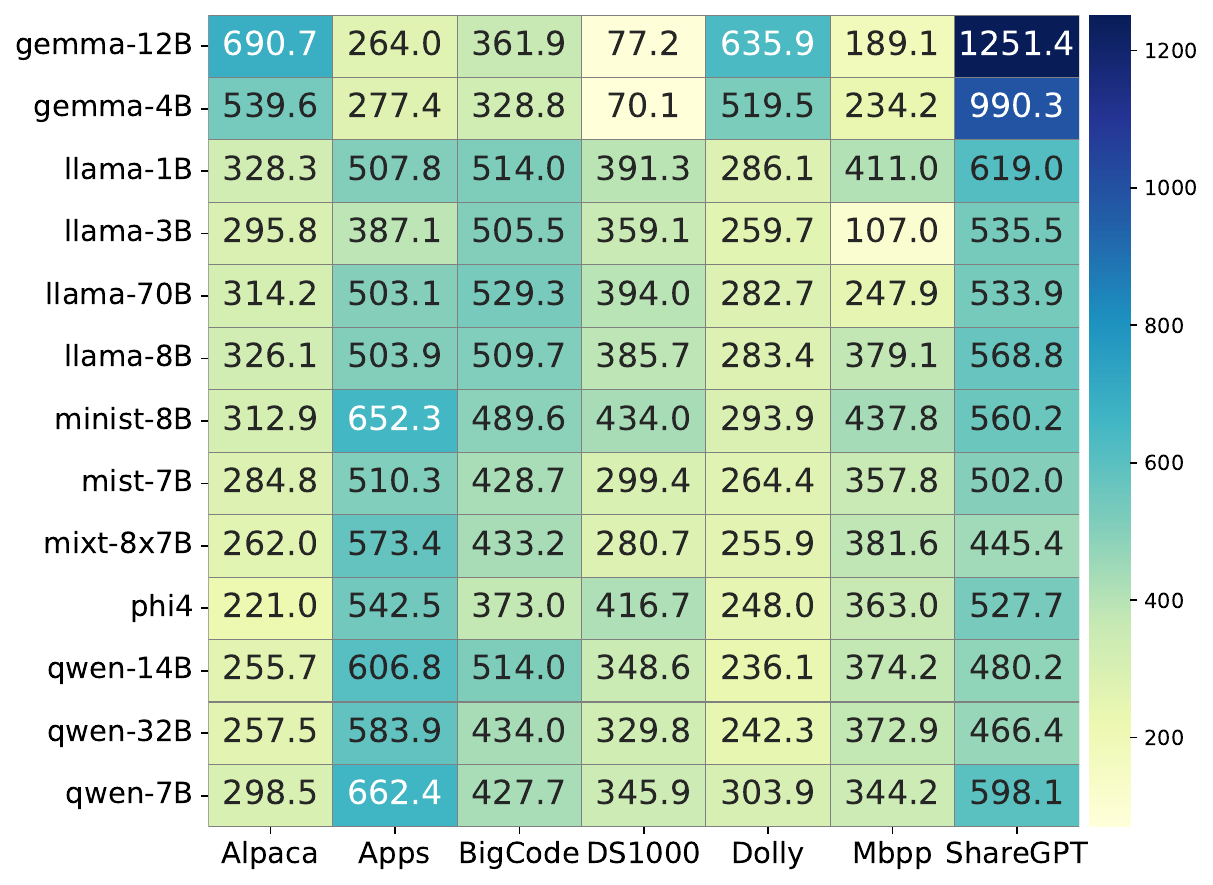}
    \caption{Average of mean response size.}
    \label{subfig:heatmap_mean}
      \end{subfigure}
      \hfill
      \begin{subfigure}[b]{0.49\textwidth}
        \includegraphics[width=\linewidth]{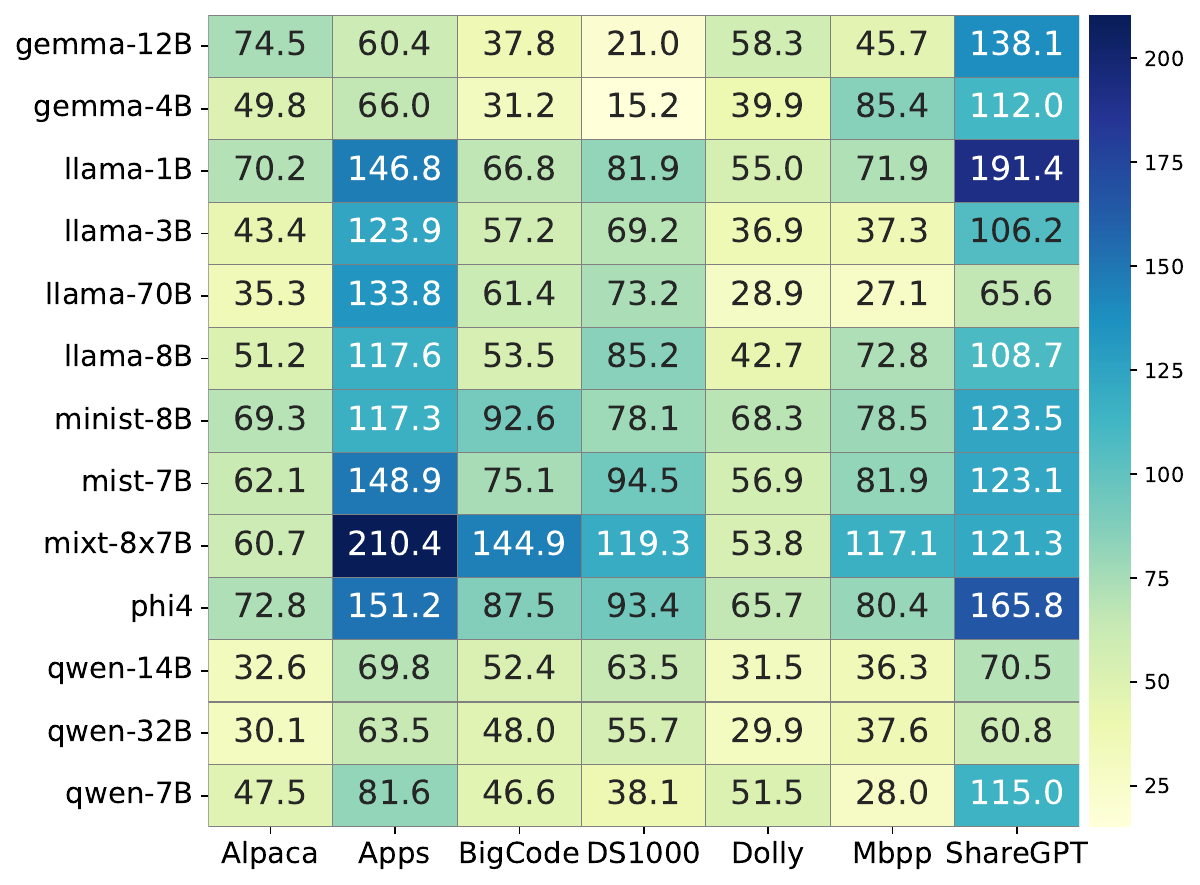}
        \caption{Average std-dev of response size.}
        \label{subfig:heatmap_std}
      \end{subfigure}
    \caption{\textbf{Both the average mean and std-dev across models and dataset can vary by several hundreds of tokens}. Heatmaps depict the average of both the mean (\ref{subfig:heatmap_mean}) and std-dev (\ref{subfig:heatmap_std}) of response lengths across models and datasets for all generations performed.}
    \label{fig:heatmaps}
\end{figure}

In this section, we analyze the response length distributions in CASTILLO to characterize the generation behavior of different LLMs across the considered datasets. Our findings confirm expected trends—such as substantial variability across models and datasets—but also uncover significant variation in response lengths for the \textit{same} prompt under fixed decoding settings. This highlights the challenges in accurately predicting response lengths. Moreover, our analysis reveals notable occurrences of text degeneration in a subset of generated completions.

\subsection{Inter- and Intra-Model Variability} 

Figure~\ref{fig:heatmaps} presents a heatmap of the average mean (\ref{subfig:heatmap_mean}) and standard deviation (\ref{subfig:heatmap_std}) of response lengths across all $\langle$model, dataset$\rangle$ combinations. 
We observe pronounced disparities in both metrics. Models consistently produce longer responses for certain datasets, notably ShareGPT and Apps~\citep{hendrycks2021apps}, with mean response lengths varying by several hundred tokens across models.
The standard deviation heatmap (Figure~\ref{subfig:heatmap_std}) further illustrates intra-model variability, with response length variation ranging from a few dozen tokens (e.g., gemma-4B on DS-1000) to a couple hundreds of tokens (e.g., mixt-8x7B). These findings emphasize the inherent stochasticity of autoregressive generation, even under fixed sampling parameters, and underscore the difficulty of response length prediction.

\begin{figure}
  \centering
  \begin{subfigure}[b]{0.98\textwidth}
    \includegraphics[width=\linewidth]{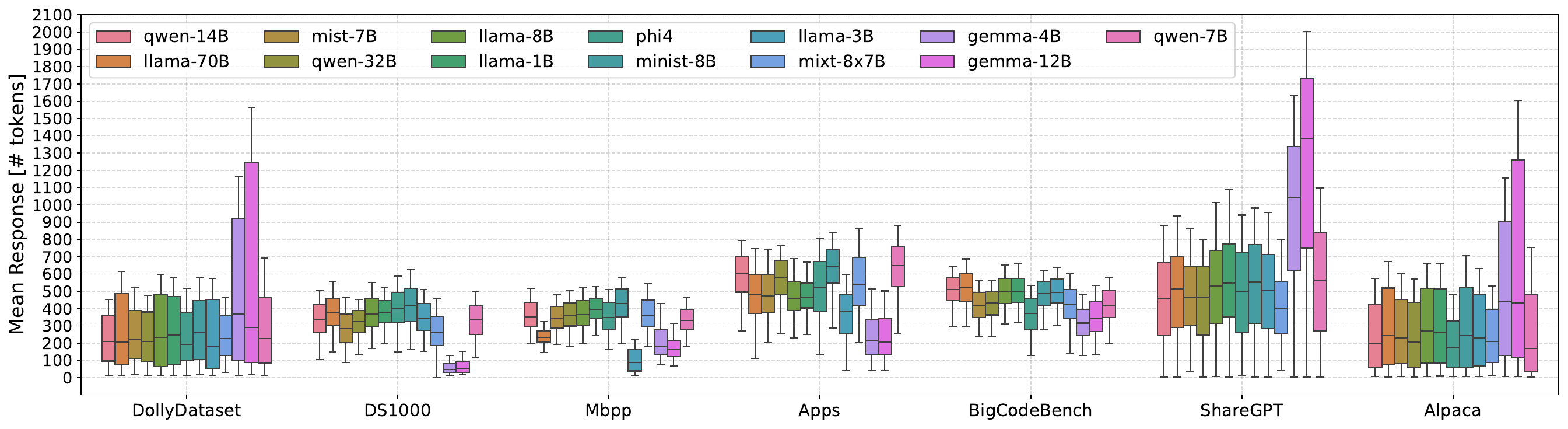}
    \caption{Mean response length distribution across models and datasets depicts significant inter-model variation in response lengths. Box extents delimit inter-quartile range, whiskers depict 1\textsuperscript{st} and 90\textsuperscript{st} percentile. }
    \label{subfig:boxplot_mean-resp-length}
  \end{subfigure}
  \vfill
  \begin{subfigure}[b]{0.98\textwidth}
    \includegraphics[width=\linewidth]{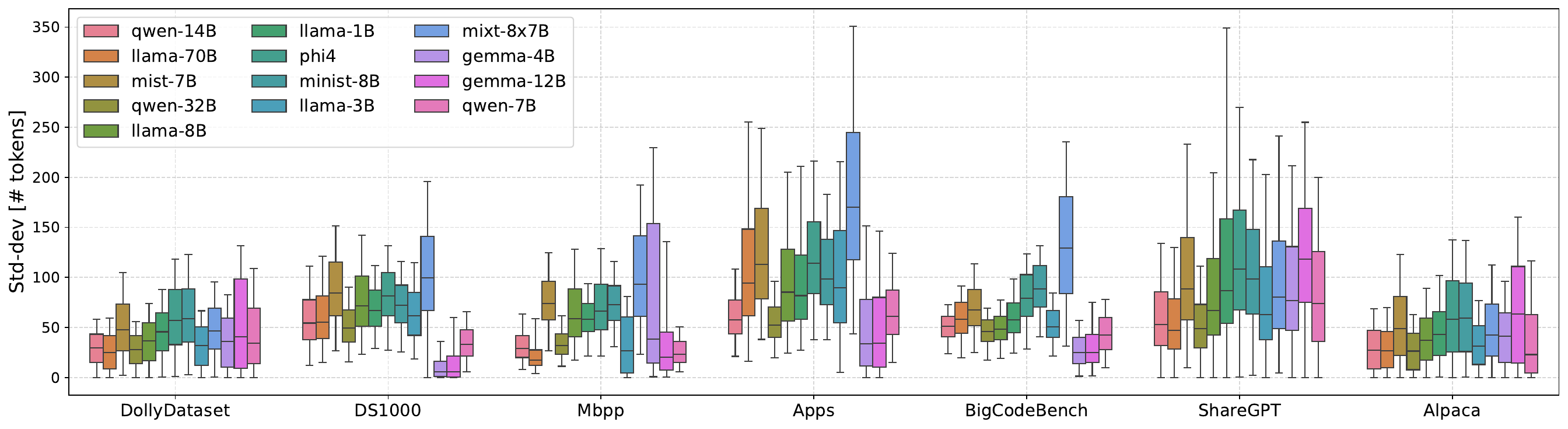}
    \caption{Distribution of standard deviation of model response lengths across datasets. Box extents delimit inter-quartile range, whiskers depict 1\textsuperscript{st} and 99\textsuperscript{st} percentile.}
    \label{subfig:boxplot_std-dev}
  \end{subfigure}
  \caption{\textbf{LLMs exhibit substantial variation in their distribution of both mean and std-dev of their responses.} Figure\ref{subfig:boxplot_mean-resp-length} shows large disparities across both models and datasets, even under identical generation settings.}
  \label{fig:box-bar_model_dataset}
\end{figure}

\paragraph{Distributions Across Models and Datasets}  

To further explore the range of variability, Figure~\ref{fig:box-bar_model_dataset} presents boxplots of the mean (\ref{subfig:boxplot_mean-resp-length}) and standard deviation (\ref{subfig:boxplot_std-dev}) of response lengths across all $\langle$prompt, model$\rangle$ pairs, grouped by dataset. The wide interquartile ranges and extended whiskers illustrate the breadth of generation behavior across LLMs.

We observe a consistent pattern in which code-related datasets (Mbpp, Apps, DS-1000, BigCodeBench) elicit longer responses compared to instruction-tuned text datasets (e.g., Alpaca, Dolly). Interestingly, the Gemma models deviate from this trend, producing longer responses on text-based datasets. This inversion highlights the importance of model-specific tuning and architectural differences in influencing output characteristics.

\begin{figure}
    \centering
    \includegraphics[width=0.98\linewidth]{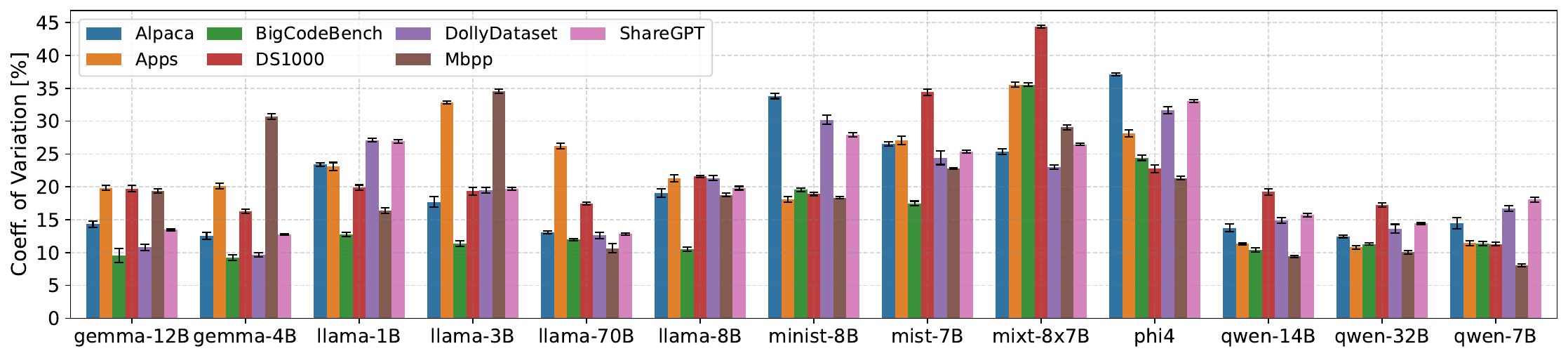}
    \caption{\textbf{High coefficient of variation between responses for the same prompt from an LLM}. Figure depicts average coefficients of variation (std-dev normalized by the mean) across datasets and models ranging from $7\%$ to up to \textbf{$45\%$}. Hence, the std-dev can reach up to half of the value of the average response length in the batch.}
    \label{fig:coeff-var}
\end{figure}

\paragraph{Within-Batch Variability and Coefficient of Variation}
To quantify the variation within each batch of 10 generated responses per ⟨prompt, model⟩ pair, we compute the average coefficient of variation (CV = std-dev/mean) per model-dataset combination. Figure~\ref{fig:coeff-var} shows CV values ranging from 7\% to 45\%, indicating substantial within-batch response length variation despite fixed decoding parameters for the batch.
High intra-batch CV values suggest that LLM outputs can be highly unpredictable for certain models and datasets, particularly for code-oriented datasets like Apps, BigCode and DS-1000, where prompt structure and complexity likely amplify generation variance. This level of unpredictability presents a challenge for inference-time schedulers and systems that rely on deterministic or narrowly distributed response length estimates.

\begin{figure}
  \centering
  \begin{subfigure}[b]{0.4\textwidth}
    \includegraphics[width=\linewidth]{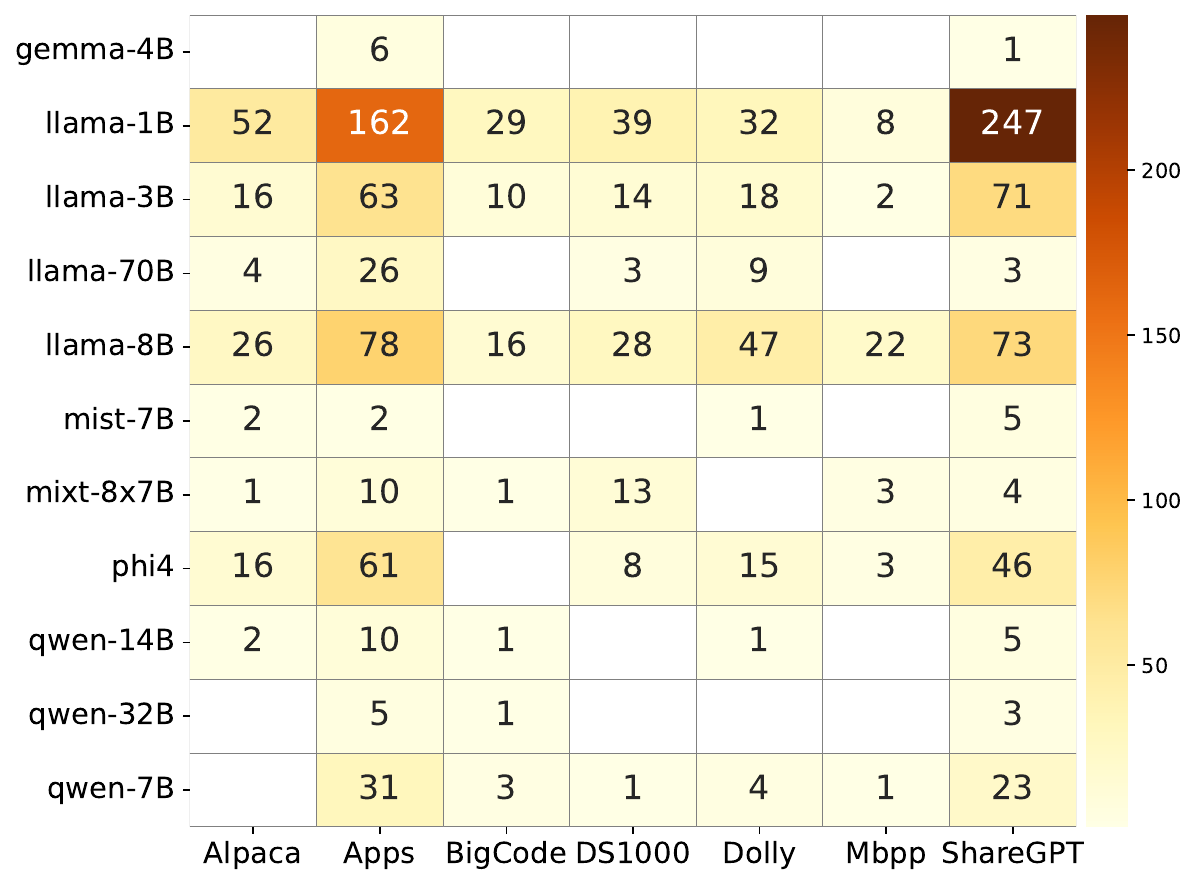}
    \caption{Number of samples suffering from test degeneration.}
    \label{subfig:heatmap-textdegen}
  \end{subfigure}
  \hfill
  \begin{subfigure}[b]{0.29\textwidth}
    \includegraphics[width=\linewidth]{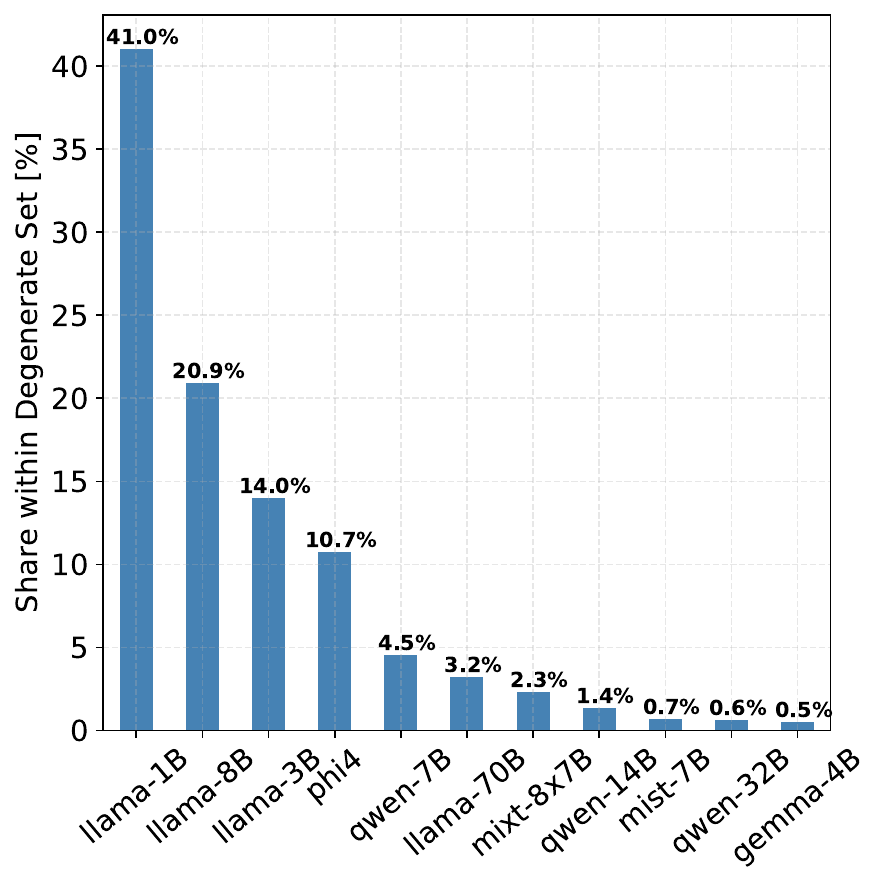}
    \caption{Share in text degeneration set by model.}
    \label{subfig:barplot-textdegen-model}
  \end{subfigure}
  \hfill
  \begin{subfigure}[b]{0.29\textwidth}
    \includegraphics[width=\linewidth]{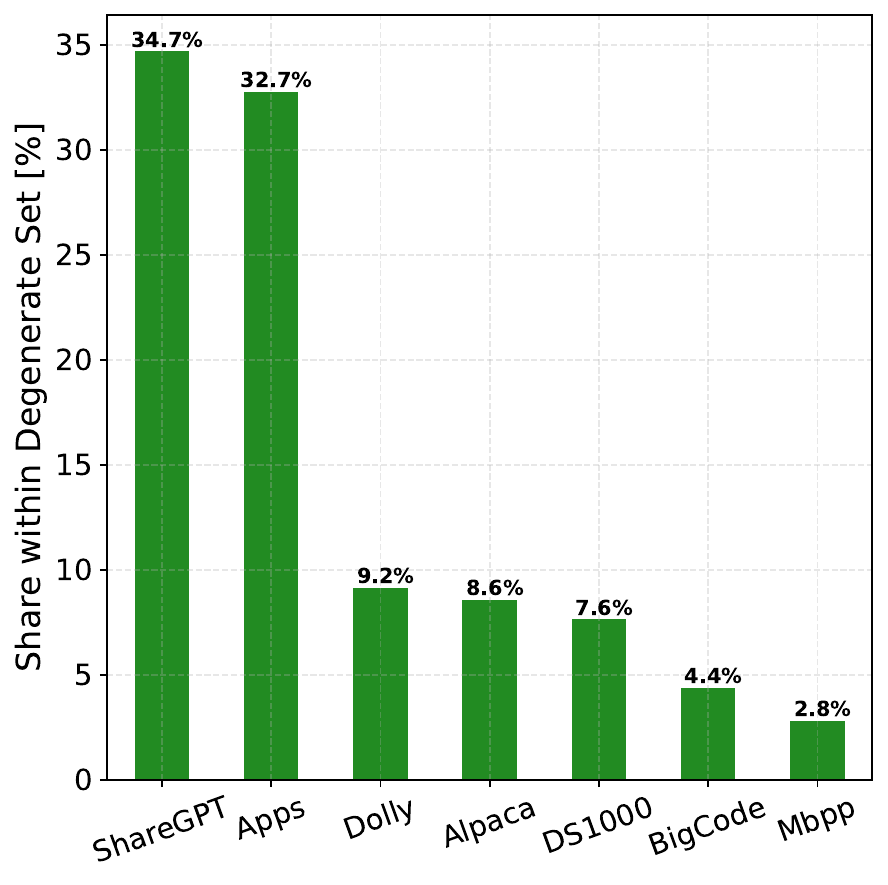}
    \caption{Share in text degeneration set by dataset.}
    \label{subfig:barplot-textdegen-dataset}
  \end{subfigure}
  \caption{\textbf{We identify a total of 956 text degeneration cases across all models and datasets.} Figure\ref{subfig:heatmap-textdegen} depicts a heatmap of the number of text degeneration cases by model and dataset. Figures~\ref{subfig:barplot-textdegen-model} and ~\ref{subfig:barplot-textdegen-dataset} show the share of samples from the total 956 according to the model and dataset, respectively.}
  \label{fig:textdegen}
\end{figure}

\subsection{Text Degeneration Cases}

We define \textit{text degeneration} as cases where generated completions exhibit artifacts such as excessive repetition, incoherence, or abrupt verbosity, often reaching the maximum token limit. Figure~\ref{appfig:text-degen-char} and~\ref{appfig:text-degen-sentence-block} in Appendix~\ref{app:textdegeneration} illustrate qualitative examples of such behavior.
CASTILLO enables the identification of degeneration cases by generating 10 independent completions per prompt under fixed decoding settings. This design facilitates benchmarking the likelihood of encountering degeneration artifacts across models and datasets.

Our degeneration detection pipeline (described in Algorithm~\ref{alg:sanitizeddataset} in Appendix~\ref{app:textdegeneration}) identified a total of 956 affected samples. Figure~\ref{fig:textdegen} provides a heatmap of degeneration counts across $\langle$model, dataset$\rangle$ combinations. Of the 13 LLMs evaluated, only two—minist-8B and gemma-12B—exhibited no degeneration cases. In contrast, three models (llama-1B, llama-3B, and llama-8B) showed degeneration across all datasets.
A clear trend emerges within model families and their sizes: smaller variants (e.g., qwen-7B, llama-1B) are more prone to degeneration than their larger counterparts. Figures~\ref{subfig:barplot-textdegen-model} and~\ref{subfig:barplot-textdegen-dataset} summarize the distribution of degeneration cases across models and datasets. Notably, llama-1B alone accounts for over 40\% of all degeneration cases, which may reflect the model's limited capacity to regulate generation under open-ended prompting. On the dataset side, ShareGPT and Apps contribute the majority of degeneration cases, likely due to their high diversity and prompt verbosity.

These findings support the inclusion of both sanitized and degeneration-only subsets in CASTILLO, enabling researchers to develop robust length predictors and to better understand model-specific vulnerabilities in generation stability.

\section{Use Cases for CASTILLO}

CASTILLO is designed to serve as a flexible and extensible benchmark for analyzing and predicting the output behavior of instruction-tuned LLMs. We outline several key use cases that highlight the dataset’s applicability in both systems and machine learning research:

\paragraph{Length Prediction for Proactive Scheduling}
In latency-sensitive LLM serving systems, the ability to predict output length prior to generation is critical for informed admission control, resource allocation, and scheduling. CASTILLO provides per-sample statistics (mean, std, percentiles) over multiple completions per prompt, which enables the training and evaluation of both regression-based and classification-based output length predictors. These models can be integrated into serving infrastructures to support proactive inference scheduling strategies.

\paragraph{Benchmarking Model-Specific Generation Behavior}
CASTILLO enables researchers to compare how different open-source LLMs respond to the same prompt distribution under identical decoding parameters. By including models across diverse architecture families and parameter scales (1B–70B), our dataset supports detailed inter- and intra-model comparisons. This facilitates analyses of length variability, degeneration tendencies, and generation style, which can inform the selection and deployment of models for specific application domains.

\paragraph{Degeneration Detection and Filtering}
The dataset includes both sanitized and degeneration-only subsets, enabling research into degeneration detection, mitigation strategies, and generation stability analysis. Degeneration patterns—such as token repetition, incoherence, or excessive verbosity—are often subtle and model-specific. Our dataset provides the necessary signals (e.g., outlier response lengths, high variance) to develop heuristics or learning-based detectors for pathological generations.

\paragraph{Downstream Integration for System Simulations}
CASTILLO can serve as a simulation backend for system-level studies in LLM inference. For example, system designers can use the statistical length distributions to emulate response lengths in multi-tenant inference clusters, or to test token-level schedulers and memory managers under realistic generation profiles. By decoupling evaluation from online generation, researchers can prototype and benchmark new scheduling policies efficiently.
\section{Limitations and Future Work}
\label{sec:limitations}

CASTILLO offers a comprehensive and empirically grounded foundation for analyzing LLM response length distributions, addressing a crucial gap in current benchmarking and systems design. However, we identify the following limitations and opportunities for future work.

\paragraph{Transformer-centric Scope}
All models used for building our dataset are based on the transformer architecture, reflecting its current dominance in LLM research and deployment. While our findings are broadly applicable, they may not generalize to alternative architectures such as state-space models~\citep{gu2022ssm, gu2023mamba} or LSTM-based systems~\citep{beck2024xlstm}. Future iterations of CASTILLO could incorporate such models to assess whether similar length variability patterns emerge beyond the transformer architecture.

\paragraph{Extensibility to More Models and Datasets}
While CASTILLO includes a diverse range of 13 instruction-tuned LLMs and 7 datasets, the long tail of model and data heterogeneity in practice suggests significant room for growth. To support community-driven expansion, we publicly release the dataset and a modular codebase (see Appendix~\ref{app:availability}) that enables seamless addition of new models, datasets, and generation settings. Notably, CASTILLO also supports optional caching of prefill-stage activations (hidden states and logits), opening new avenues for analysis of how input representations correlate with output length. We envision future work could systematically explore this relationship.

\paragraph{Chain-of-Thought Specific Extensions}
Out dataset generation framework can be extended to accommodate reasoning-focused models that follow chain-of-thought framework~\citep{wei2022cot, yao2023tree}. In such settings, we propose partitioning the response into two measurable segments: (1) the reasoning trace delimited by \texttt{<think>} tokens and (2) the final answer. This would enable fine-grained prediction of intermediate vs. final response lengths, and support a deeper understanding of how models allocate tokens to reasoning versus conclusions. Given the demonstrated model-specific variation in generative behavior, we hypothesize that reasoning lengths may also reflect distinct architectural or model "personalities".

\paragraph{Generation-Configuration Interactions}
While CASTILLO generates responses assuming the respective model's default decoding settings, generation configuration remains a powerful lever that modulates output variability. Our dataset includes metadata for temperature, top-k, and top-p settings, and can be extended to systematically study their interactions with both prompt structure and model identity. Future work could incorporate model-specific controlled ablation studies to quantify the marginal effect of each decoding parameter on response length distribution and degeneration likelihood. This process is model-dependent, and outside of our scope of inter-model variability. 
\section{Conclusion}

We introduced CASTILLO, a large-scale dataset designed to empirically characterize response length distributions across open-source instruction-tuned LLMs. By systematically measuring response length variability across 13 models and 7 diverse datasets, it provides a rich foundation for building predictive models that support proactive inference scheduling. Beyond its utility for systems research, our dataset enables the investigation of model-specific generation behaviors, including intra-model variance and partial text degeneration phenomena. We release the dataset, codebase, and baseline tools to encourage reproducibility and community-driven extensions, and we envision CASTILLO as a versatile tool for research at the intersection of machine learning and systems.




\bibliographystyle{abbrvnat}
\bibliography{refs}


\appendix

\section{Dataset and Code Availability}
\label{app:availability}

We make our dataset and code publicly available. 
We uploaded our dataset to the HuggingFace platform at \url{https://huggingface.co/datasets/danfperam/castillo} under the Creative Commons Attribution License (CC BY 4.0).
Moreover, we release the documented code used for generating the dataset, as well the the code required for running the experiments in \url{https://github.com/DanielFPerez/castillo} under the Apache 2.0 license.


\section{Analyzing and Processing the Source Datasets}
\label{app:sourcedatasets}
We refer to "Source Datasets" as the text corpora that we use as source to generate the prompts to the models.
We focus on open source instruction-following data corpora and prioritize those that have been generated by "natural" model-user interactions, and those that focus on code generation. 
For all text corpora, we split each respective dataset with a 70\%-20\%-10\% training-validation-test set splits.
We intend for future works that the train and validation sets serve develop length predictor models. The test set will be a hold-out dataset that will subsequently be used for the scheduler experiments in systems research, in which the output length predictor will be deployed.


\subsection{Dolly Dataset}

Dataset presented by \cite{DatabricksBlog2023DollyV2} consisting of 15k intruction-following records of Databricks employees with their model. We access the dataset from \url{https://huggingface.co/datasets/databricks/databricks-dolly-15k}.

\paragraph{Schema} The schema of each sample of the dataset is a dictionary with the following fields:
\begin{small}
\begin{verbatim}
{   'instruction': 'Given a reference text about Lollapalooza, where does ...',
    'context': 'Lollapalooza is an annual American festival...',
    'response': 'Lollapalooze is an annual musical festival held ...',
    'category': 'closed_qa'
}
\end{verbatim}
\end{small}

The \verb|context| field is used by some samples as additional information when prompting the model, while the \verb|category| field contains a label classifying the question into eight categories: \verb|brainstorming|, \verb|classification|, \verb|closed_qa|, \verb|creative_writing|, \verb|general_qa|, \verb|information_extraction|, \verb|open_qa|, \verb|summarization|.

\begin{figure}[ht]
  \centering
  \begin{subfigure}[b]{0.95\textwidth}
    \includegraphics[width=\linewidth]{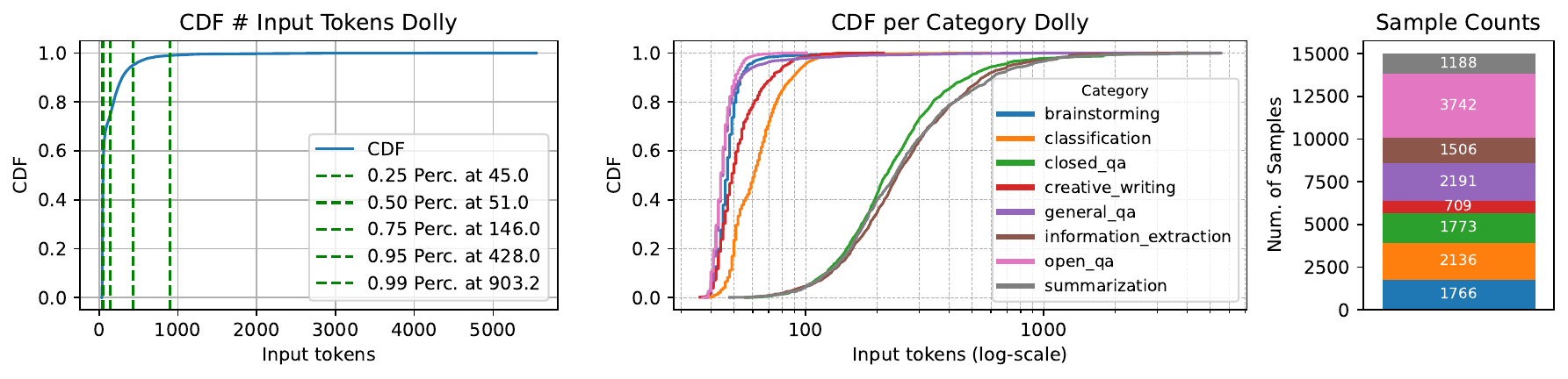}
    \caption{DollyDataset: distributions over the complete 15000 samples.}
    \label{fig:sub1}
  \end{subfigure}
  \vfill
  \begin{subfigure}[b]{0.95\textwidth}
    \includegraphics[width=\linewidth]{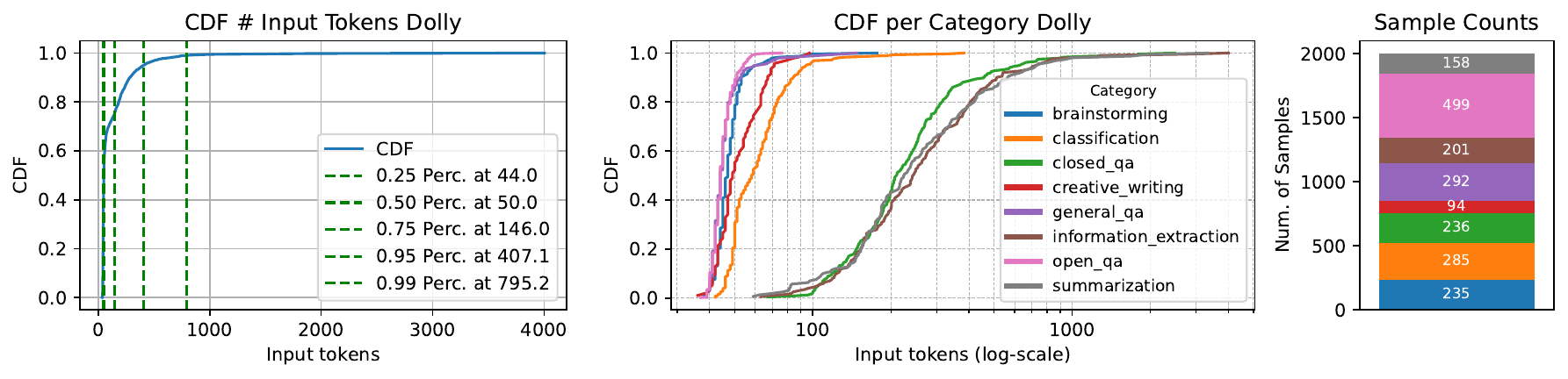}
    \caption{DollyDataset: distribution over the selected 2000 samples.}
    \label{fig:sub2}
  \end{subfigure}
  \caption{\textbf{DollyDataset} prompt lengths distributions (number of tokens) from both the complete dataset, and from the 2000 randomly stratified sampled instances. The sampled subset shares a similar distribution as the original dataset.}
  \label{appfig:dollydist}
\end{figure}

Figure~\ref{appfig:dollydist} shows the \ac{CDF} of the tokenized input for all prompts in the dolly dataset using the Llama-3.2-3B model. The right-hand plot shows also the eCDF of each of the categories.

\paragraph{Text prompt} We use the following prompt structure for each sample of the DollyDataset:
\begin{small}
\begin{verbatim}
    sample['instruction'] + "\nContext: " + sample['context']
\end{verbatim}
\end{small}
where the "Context" line is added only for samples that have a \verb|context| field, and omitted otherwise. 

\paragraph{Stratified Split} The DollyDataset includes a categorical variable indicating the type of question or task (the \verb|category| feature). To maintain consistency and ensure proper statistical data handling, we perform stratified sampling according to the \verb|category| field \textit{both} for randomly selecting the initial 2000 samples from the dataset, and also for the subsequent train-validation-test splits.

\subsection{ShareGPT dataset}
The ShareGPT dataset is a large, publicly released collection of anonymized user–ChatGPT conversational exchanges scraped from publicly shared chat links, comprising paired prompts and AI responses for training and evaluating dialogue models~\citep{chiang2023vicuna}. We choose the following huggingface version of the ShareGPT dataset \url{https://huggingface.co/datasets/anon8231489123/ShareGPT_Vicuna_unfiltered}.

\paragraph{Schema} The \verb|json| file that corresponds to the data is a list of \verb|dict| elements. Each of these elements has the following structure:
\begin{small}
\begin{verbatim}
{   'id': 'xd92L6L_48',
    'conversations': [ {'from': 'human', 'value': 'Tell me a story.'},
                       {'from': 'gpt', 'value': 'Once upon a time ....'},
                       {'from': 'human', 'value': 'Make it shorter, '}, ...] }
\end{verbatim}
\end{small}

\paragraph{Dataset cleansing}. The original dataset has 94145 samples. First, we remove 550 data samples from the original dataset as we identify that they do not contain any conversations. Moreover, we notice that some original long conversations interactions with ChatGPT were split into multiple data samples, which causes some data samples to start their conversation with the \verb|gpt| input. We discard those data samples as well, since we do not want to bias the response from the considered \acp{LLM} by CASTILLO with ChatGPT responses. Future work could consider caching the results from the CASTILLO models and feeding the continuation of the ShareGPT conversations that start with \verb|human|.
After considering only data samples starting with \verb|human|, the dataset results in 58827 data samples. 

\paragraph{Text prompt} We use the following prompt structure for each sample of the ShareGPT dataset:
\begin{small}
\begin{verbatim}
    sample['conversations'][0]['value']
\end{verbatim}
\end{small}

\begin{figure}[ht]
  \centering
  \begin{subfigure}[b]{0.95\textwidth}
    \includegraphics[width=\linewidth]{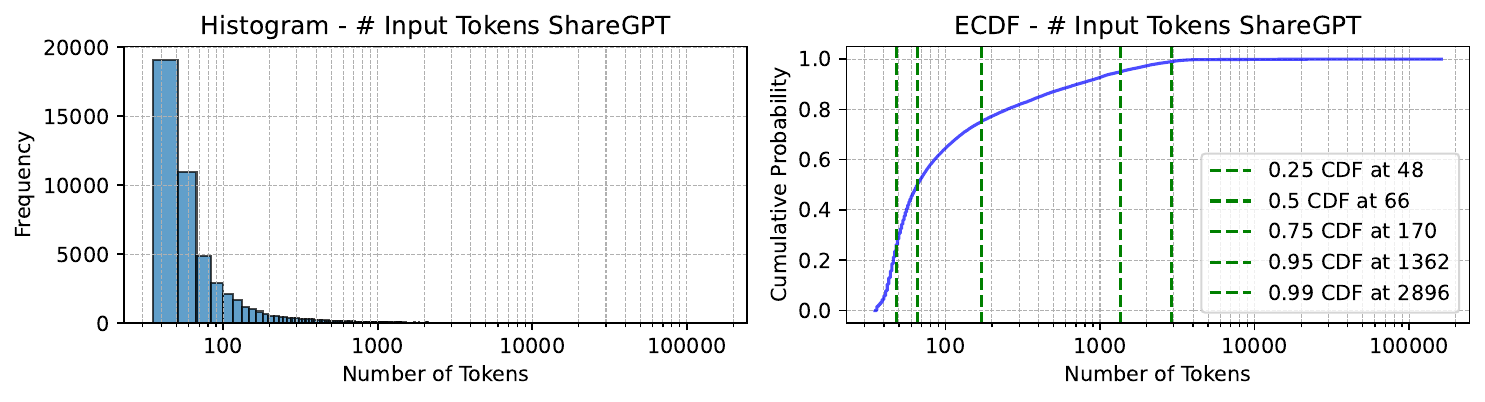}
    \caption{ShareGPT: distributions over the filtered 58827 samples.}
    \label{appfig:sharegptdist-all}
  \end{subfigure}
  \vfill
  \begin{subfigure}[b]{0.95\textwidth}
    \includegraphics[width=\linewidth]{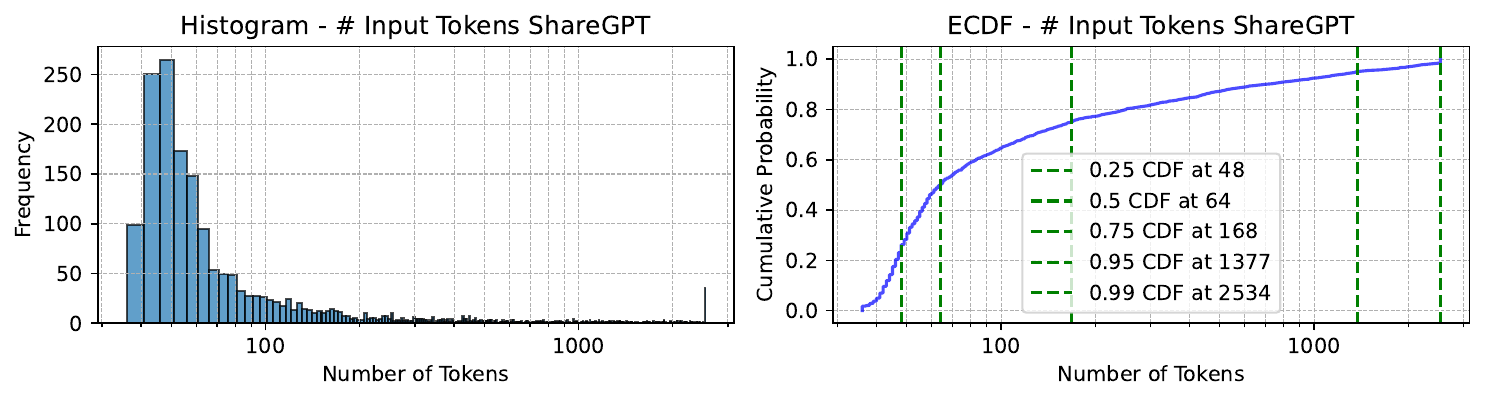}
    \caption{ShareGPT: distribution over the selected 2000 samples.}
    \label{appfig:sharegptdist-2k}
  \end{subfigure}
  \caption{\textbf{ShareGPT} prompt lengths distributions (number of tokens) from both the complete dataset, and from the 2000 randomly sampled instances. The sampled subset shares a similar distribution as the original dataset.}
  \label{appfig:sharegptdist}
\end{figure}

Figure~\ref{appfig:sharegptdist} shows the distribution of the length of the tokenized user prompts in terms of number of tokens for both the filtered version of the dataset and for the 2000 samples considered for CASTILLO. Bpoth distributions are similar, with closely aligned percentiles values. It shows how 50\% of the ShareGPT prompts are less or equal to $\sim$65 tokens long, and 95\% are equal or shorter than 1400. 
Finally, there are some prompts that are excessively long in the filtered 58k samples (see long tail in Figure~\ref{appfig:sharegptdist-all}), with a maximum length of 161306 tokens, far beyond the context window of most open source \acp{LLM}. 
From the selected 2000 samples, we truncate their length to a maximum of 2500 tokens, given that the 99\textsuperscript{th} percentile is already at 2500.

\subsection{Alpaca Dataset}
The Alpaca dataset presented by~\cite{taori2023alpaca} is a collection of 52000 instruction-following demonstrations generated using self-instruct techniques, where prompts are crafted and outputs are synthesized by GPT-3 to fine-tune smaller language models like LLaMA.  We choose the following huggingface version of the Alpaca dataset that already performs some cleaning of the original dataset \url{https://huggingface.co/datasets/yahma/alpaca-cleaned}.

\paragraph{Schema} The Alpaca dataset is provided as a list of \verb|dict| elements, where aech sample of the dataset has the following structure:
\begin{small}
\begin{verbatim}
{   'instruction': 'Explain why the following fraction is equivalent to 1/4',
    'input': '4/16',
    'output': 'The fraction 4/16 is equivalent ....'
}
\end{verbatim}
\end{small}
The \verb|input| field is used by some samples as additional information when prompting the model.

\paragraph{Text prompt} We use the following prompt structure for each sample of the Alpaca dataset:
\begin{small}
\begin{verbatim}
    sample['instruction'] + "\nInput: " + sample['context']
\end{verbatim}
\end{small}
Analogous to the DollyDataset, the "Input" line is added only for samples that have a \verb|input| field, and omitted otherwise.

\begin{figure}[ht]
  \centering
  \begin{subfigure}[b]{0.95\textwidth}
    \includegraphics[width=\linewidth]{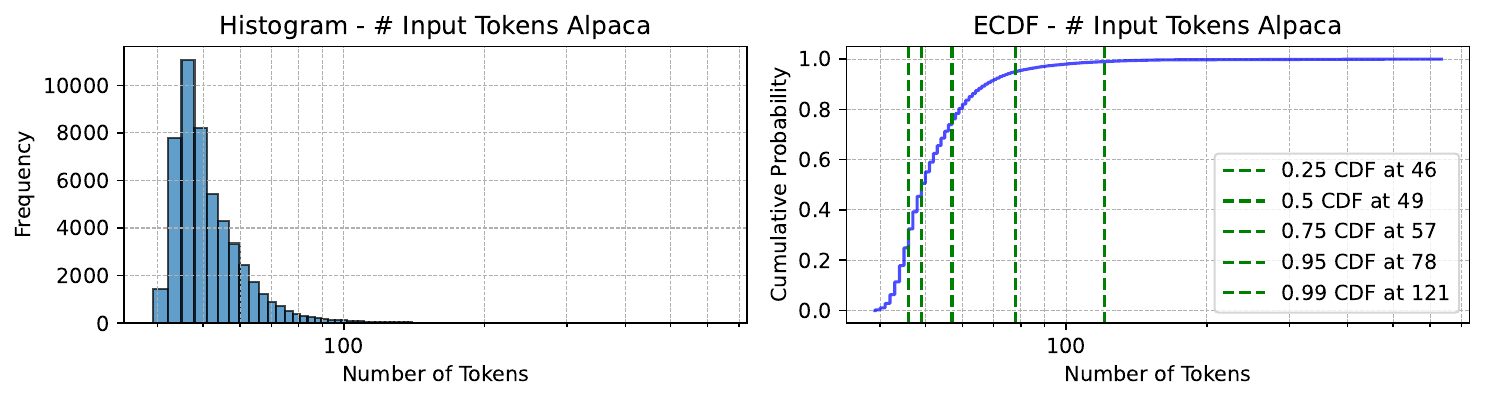}
    \caption{Alpaca: distributions over the filtered 51760 samples.}
    \label{appfig:alpacadist-all}
  \end{subfigure}
  \vfill
  \begin{subfigure}[b]{0.95\textwidth}
    \includegraphics[width=\linewidth]{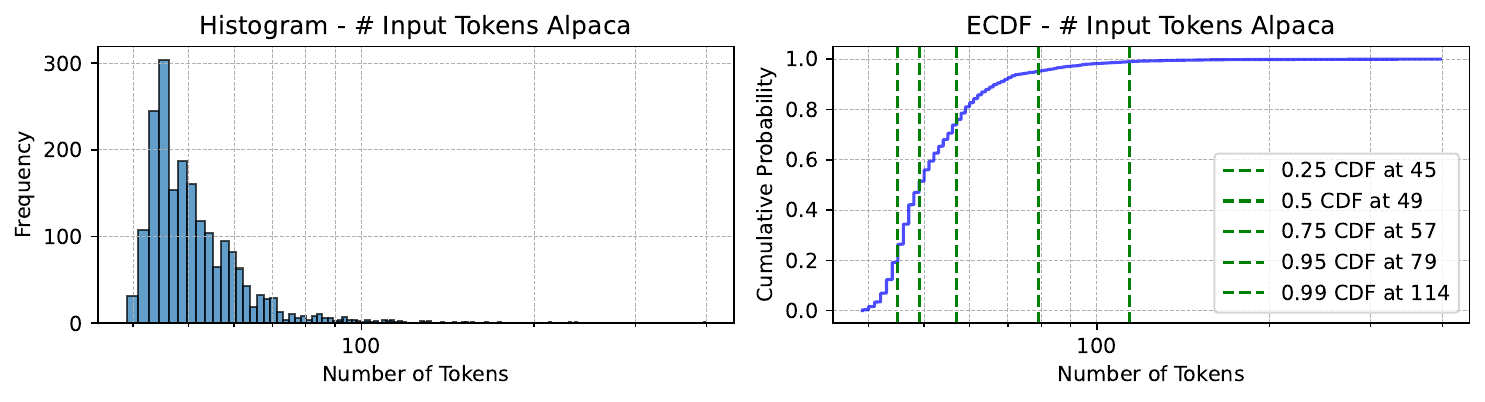}
    \caption{ShareGPT: distribution over the selected 2000 samples.}
    \label{appfig:alpacaptdist-2k}
  \end{subfigure}
  \caption{\textbf{Alpaca} prompt lengths distributions (number of tokens) from both the complete dataset, and from the 2000 randomly sampled instances. The sampled subset shares a similar distribution as the original dataset.}
  \label{appfig:alpacadist}
\end{figure}

Figure~\ref{appfig:alpacadist} shows the distribution of the length of the resulting input prompts using the Llama-3.2-1B tokenizer. This dataset has a shorter tail than the previous considered. The longest input prompt recorded is 630 tokens long, and 99\% of the input prompts are shorter than 114 tokens. Moreover, Figure~\ref{appfig:alpacaptdist-2k} shows a similar distribution as the original 51760 samples form Figure~\ref{appfig:alpacadist-all}. 

\subsection{Mbpp Dataset}

The Mbpp (Mostly Basic Python Problems) dataset was originally published by ~\cite{austin2021mbpp}, and we access the dataset from \url{https://github.com/google-research/google-research/tree/master/mbpp}. The original dataset consists of 1000 crowd-sourced python programming problems. We choose to take the subset of the data that is hand-verified by the authors, which they called a \textit{sanitized} version, comprising of 974 crowd-sourced Python programming tasks designed to be solvable by entry-level programmers.

\paragraph{Schema} Each sample in the Mbpp dataset has the following structure: 
\begin{small}
\begin{verbatim}
{'source_file': 'Benchmark Questions Verification V2.ipynb',
 'task_id': 3,
 'prompt': 'Write a python function to identify non-prime numbers.',
 'code': 'import math\ndef is_not_prime(n) ... ',
 'test_imports': [],
 'test_list': ['assert is_not_prime(2) == False',
  'assert is_not_prime(10) == True'
}
\end{verbatim}
\end{small}

\paragraph{Text prompt} We follow the prompt format described by \citet{austin2021mbpp}:
\begin{small}
\begin{verbatim}
   "You are an expert Python programmer, and here is your task: " 
   + sample['prompt'] + 
   "Your code should pass these tests:\n\n" 
   +"\n".join(sample['test_list']) + "\n"
\end{verbatim}
\end{small}

\begin{figure}
    \centering
    \includegraphics[width=0.95\linewidth]{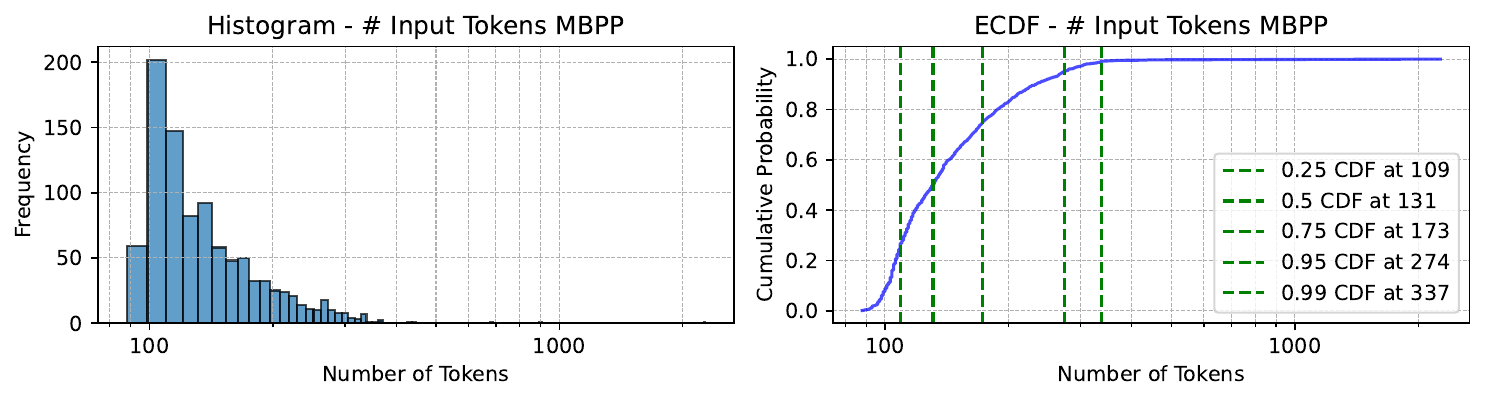}
    \caption{\textbf{Mbpp} distribution of the length of prompts in terms of number of tokens over the 974 samples.}
    \label{appfig:mbbpdist}
\end{figure}

Figure~\ref{appfig:mbbpdist} depicts the distribution of the length of the tokenized user prompts in terms of number of tokens.
The input prompts in the dataset have an average length of approximately 154 tokens after tokenization, with lengths ranging from 88 to 2,265 tokens. The distribution is moderately skewed, with 75\% of prompts shorter than 173 tokens and only 1\% exceeding 337 tokens.

\subsection{Apps Dataset}

The Apps (Automated Programming Progress Standard) dataset, introduced by \cite{hendrycks2021apps}, is a benchmark designed to evaluate the code generation capabilities of \acp{LLM}. It comprises 10,000 Python programming problems sourced from open-access coding platforms like Codeforces and Kattis, with difficulties ranging from introductory to competitive programming levels. We access the dataset from \url{https://huggingface.co/datasets/codeparrot/apps}, and choose the "train" split, which contains 5000 samples.

\paragraph{Schema} Each sample in the Apps dataset has the following structure:
\begin{small}
\begin{verbatim}
{'problem_id': 899,
 'question': 'Chef wants to teach a lesson of sharing to the students. \nTh[...]',
 'solutions': '["from math import ceil\\n\\nfor _ in range(int([...]", 
                "import math\\nT=int(input())\\n\\nfor _ in rang[...]", ...]',
 'input_output': '{"inputs": [["2", "5", "1 2 3 4 5", "5", "5 4 3 2 1"]], 
                   "outputs": [["3", "5"]]}',
 'difficulty': 'interview',
 'url': 'https://www.codechef.com/CROS2020/problems/COCR106',
 'starter_code': ''}
}
\end{verbatim}
\end{small}

\paragraph{Text prompt} We follow the prompt format described by \citet{hendrycks2021apps}:
\begin{small}
\begin{verbatim}
   "QUESTION: " + sample["question"] + "\n\n" +
   "Your code should start with the following: \n" + 
   sample['starter_code'] + "\n" +
   "The following is the standard input and output format: \n" +
   "Inputs:\n" + 
   "\n".join([str(elem) for elem in json.loads(sample['input_output']['inputs'])]) +
   "\nOutputs:\n"
   "\n".join([str(elem) for elem in json.loads(sample['input_output']['outputs'])])
\end{verbatim}
\end{small}
Analogous to the previous datasets, we include the rows in the text prompt for those sample feature elements for which there is actually data, e.g., we only include \verb|starter_code| if the element is non-empty.

\paragraph{Dataset cleansing}
Our analysis found a sample instances that relates to factorial calculations, which yields a very long element in the \verb|datasample['input\_output']['output']`| list that prevents it from being transformed to a Dict. 
Additionally, after tokenizing all the samples of the dataset, we found instances whose tokenization reaches millions of tokens (maximum value 10.489.768).
Without loss of generality, we remove these samples from the dataset before performing the splits. Further analysis revelas that the 99\textsuperscript{th} percentile of the tokenized text prompts lies at 2422, so we truncate all prompts larger than 2500 tokens.

\paragraph{Stratified Split} The Apps dataset includes a \verb|difficulty| feature per sample, indicating the difficulty of programming task: \verb|introductory|, \verb|interview|, and \verb|competition|. To maintain consistency and ensure proper statistical data handling, we perform stratified sampling according to the \verb|difficulty| variable \textit{both} for randomly selecting the initial 2000 samples from the dataset, and also for the subsequent train-validation-test splits.

\begin{figure}[ht]
  \centering
  \begin{subfigure}[b]{0.95\textwidth}
    \includegraphics[width=\linewidth]{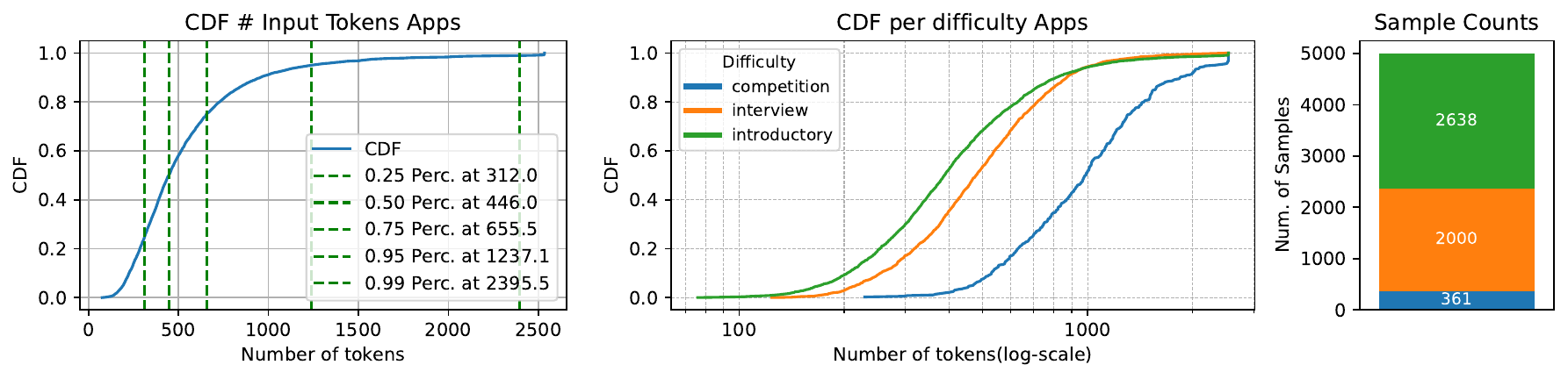}
    \caption{Apps dataset: distributions over the complete 4999 samples.}
    \label{fig:sub1}
  \end{subfigure}
  \vfill
  \begin{subfigure}[b]{0.95\textwidth}
    \includegraphics[width=\linewidth]{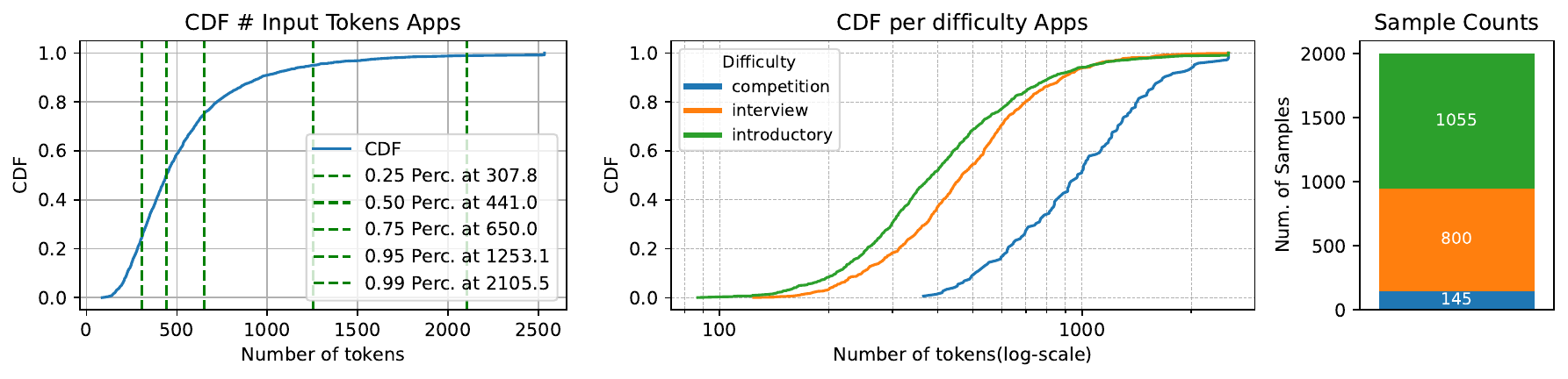}
    \caption{Apps dataset: distribution over the selected 2000 samples.}
    \label{fig:sub2}
  \end{subfigure}
  \caption{\textbf{Apps} dataset prompt lengths distributions (number of tokens) from both the complete dataset, and from the 2000 randomly stratified sampled instances. The sampled subset shares a similar distribution as the original dataset.}
  \label{appfig:appsdist}
\end{figure}

Figure~\ref{appfig:appsdist} shows the distribution of the length of the tokenized user prompts in terms of number of tokens for both the filtered version of the dataset and for the 2000 samples considered.
The input prompts in the APPS dataset have a mean token length of approximately 545, with a wide range spanning from 87 to 2,534 tokens. The distribution is heavily right-skewed, with 75\% of prompts under 650 tokens and the top 1\% exceeding 2,100 tokens. As shown in Figure~\ref{appfig:appsdist}, the prompt length correlates with problem difficulty, with competition-level tasks tending to have significantly longer inputs than interview or introductory problems. The dataset is dominated by introductory and interview tasks, comprising over 90\% of the total sample count.

\subsection{DS-1000 Dataset}

The DS-1000 dataset, introduced by~\citet{lai2023ds1000}, is a benchmark comprising 1,000 real-world data science coding problems collected from StackOverflow, covering seven widely-used Python libraries such as NumPy, Pandas, and Matplotlib. To prevent memorization by models, many problems have been perturbed through surface or semantic modifications, ensuring they differ from their original StackOverflow sources. We access the dataset from \url{https://huggingface.co/datasets/xlangai/DS-1000}. 

\paragraph{Schema} Each sample in the DS-1000 dataset has the following structure:
\begin{small}
\begin{verbatim}
{'prompt': "Problem:\nIn pandas, how do I replace &AMP; with '&'[...]",
 'reference_code': "def g(df):\n    return df.replace('&AMP;','&', rege[...]",
 'metadata': {'problem_id': 100, 
              'library_problem_id': 100, [...]},
 'code_context': 'import pandas as pd\nimport numpy as np\nimport copy\n[...]'
}
\end{verbatim}
\end{small}

\paragraph{Text prompt} According to \citet{lai2023ds1000}, we take the \verb|prompt| feature from each sample as its text prompt. As shown in the huggingface-hosted DS-1000 dataset, this feature already contains all the information necesary for building the prompt to the model.

\begin{figure}
    \centering
    \includegraphics[width=0.95\linewidth]{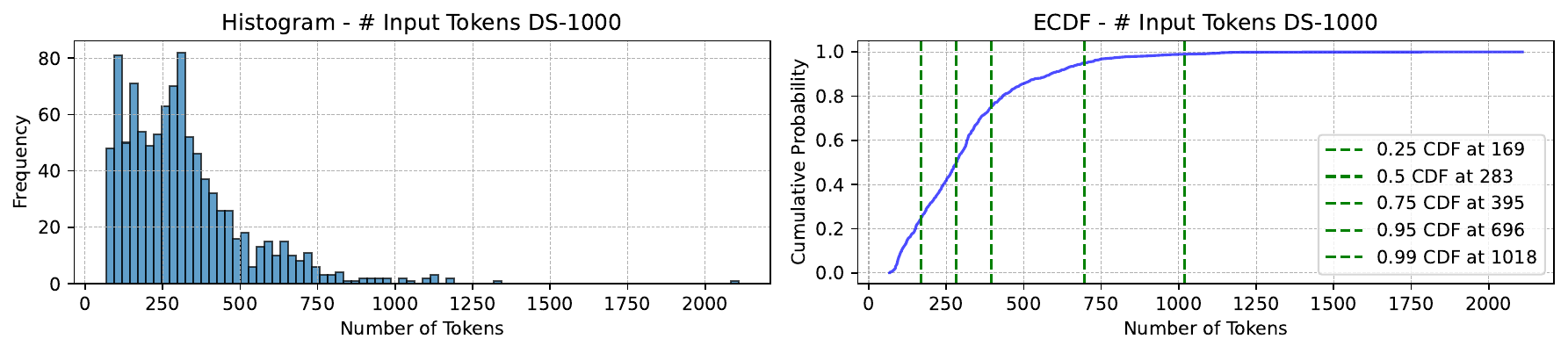}
    \caption{\textbf{DS-1000} distribution of the length of prompts in terms of number of tokens over the 1000 samples.}
    \label{appfig:ds1000pdist}
\end{figure}

The input prompts in the DS-1000 dataset have a mean token length of around 317, with values ranging from 67 to 2,109 tokens. Figure~\ref{appfig:ds1000pdist} depicts that the distribution of input length tokens is right-skewed, as 75\% of prompts are shorter than 395 tokens, while only 1\% exceed 1,018 tokens. The histogram reveals a concentration of prompts between 100 and 400 tokens, and the CDF plot confirms that most examples fall well below 700 tokens.

\subsection{BigCodeBench Dataset}

\citet{zhuo2025bigcodebench} introduced the BigCodeBench dataset, a comprehensive benchmark designed to evaluate \acp{LLM} on complex code generation tasks. We select the  "Instruct" variant of the dataset, which offers concise natural language prompts to test instruction-following capabilities of the models. We access the dataset through huggingface under \url{https://huggingface.co/datasets/bigcode/bigcodebench}, and we select the latest version available at the time of writing (v0.1.4), which contains 1140 samples.

\paragraph{Schema}  Each sample in the BigCodeBench dataset has the following structure:
\begin{small}
\begin{verbatim}
{'task_id': 'BigCodeBench/222',
 'complete_prompt': 'import math\nimport numpy as np\nimport matpl[...]',
 'instruct_prompt': 'Sort the given list in ascending order based on the deg[...]',
 'canonical_solution': '    sorted_list = sorted(list_input, key=l[...]',
 'code_prompt': 'import math\nimport numpy as np\nimport matplotlib.pyplot [...]',
 'test': "import unittest\nimport doctest\nclass TestCases(unittest.TestCase):\n[...]",
 'entry_point': 'task_func',
 'doc_struct': '{"description": ["Sort the given list in ascending [...]."], 
                 "notes": [], "params": ["list_input (list): The list to be sorted."], 
                 "returns": ["tuple: A tuple containing:[...]", [...]]}',
 'libs': "['math', 'numpy', 'matplotlib']"
 }
\end{verbatim}
\end{small}

\paragraph{Text prompt} \citet{zhuo2025bigcodebench} have transformed the tasks in the prompts of the dataset via pre-defined rules to create a natural-language oriented instructions for prompting \acp{LLM}. This transformation into natural language is encapsulated in the \verb|instruct_prompt| feature of the dataset samples, and we consider this string as the text prompt for our models. 

\begin{figure}
    \centering
    \includegraphics[width=0.95\linewidth]{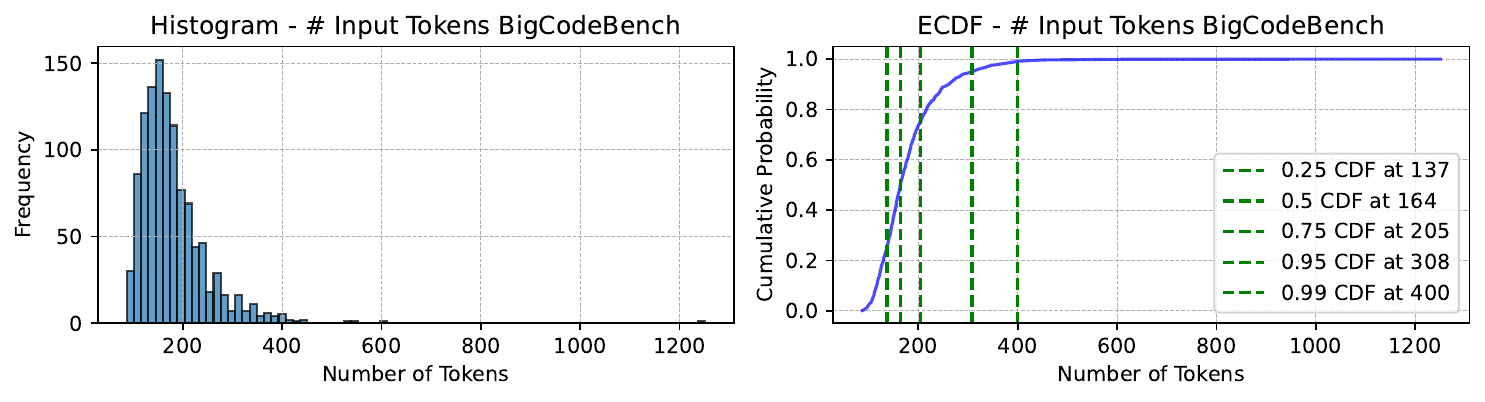}
    \caption{\textbf{BigCodeBench} distribution of the length of prompts in terms of number of tokens over the 1140 samples.}
    \label{appfig:bigcodepdist}
\end{figure}

Most input prompts in BigCodeBench are relatively concise, with a mean length of about 180 tokens and a median of 164. The distribution is tightly packed, as 75\% of prompts fall below 205 tokens and nearly all (99\%) are under 400. The histogram reveals a strong peak between 130–200 tokens, while the ECDF confirms a steep accumulation of samples in this lower range. Only a few outliers exceed 1,000 tokens, suggesting limited variation in prompt length.

\begin{table}
  \caption{Open source \acp{LLM} models considered, with their corresponding model ID from the Huggingface platform, and the number of Nvidia H100 GPUs used for generating the model responses.}
  \label{tab:llms}
  \centering
  \begin{tabular}{llcl}
    \toprule
    Short Name     & HuggingFace Model Name       &  H100s used  & Organization \\
    \midrule
    llama-1B         & meta-llama/Llama-3.2-1B-Instruct & 2 & \multirow{4}{*}{Meta}   \\
    llama-3B         & meta-llama/Llama-3.2-3B-Instruct &  2 & \\
    llama-8B         & meta-llama/Llama-3.1-8B-Instruct  &  2 &\\
    llama-70B        & meta-llama/Llama-3.3-70B-Instruct  & 4 &\\
    \addlinespace
    mist-7B       & mistralai/Mistral-7B-Instruct-v0.3 &  2 &\multirow{3}{*}{Mistral AI} \\
    minist-8B     & mistralai/Ministral-8B-Instruct-2410 &  2 &\\
    mixt-8x7B     & mistralai/Mixtral-8x7B-Instruct-v0.1 & 4 & \\
    \addlinespace
    qwen-7B       & Qwen/Qwen2.5-7B-Instruct-1M &  2 &\multirow{3}{*}{Alibaba Cloud} \\
    qwen-14B      & Qwen/Qwen2.5-14B-Instruct   &  4 &\\
    qwen-32B      & Qwen/Qwen2.5-32B-Instruct   &  4 &\\
    \addlinespace
    phi4          & microsoft/Phi-4-mini-instruct &  2 &Microsoft \\
    \addlinespace
    gemma-4B      & google/gemma-3-4b-it &  2 &\multirow{2}{*}{Google} \\
    gemma-12B     & google/gemma-3-12b-it &  2 &\\
    
    \bottomrule
  \end{tabular}
\end{table}

\section{\acp{LLM} Used}
\label{app:llmsused}

The list of the \acp{LLM} used for generating the CASTILLO dataset are listed in Table~\ref{tab:llms}. All models were instantiated using HuggingFace transformers library~\citep{wolf2020huggingface}.
For hardware resources, we generated the responses in an HPC cluster using different numnber of Nvidia H100 GPUs per model, as observed in Table 1. We allocate 20 CPUs and 64 - 128 GB of RAM per GPU utilized.


\section{Identifying and Filtering out Text Degeneration Instances}
\label{app:textdegeneration}

\begin{algorithm}[ht!]
\caption{SanitizeDataset to Identify and Sanitize Degenerated Samples}
\label{alg:sanitizeddataset}
\begin{algorithmic}[1]
\Procedure{SanitizeDataset}{dataset}
    \State Initialize \textit{degeneration\_dataset} $\gets$ [ ]
    \State Initialize \textit{sanitized\_dataset} $\gets$ [ ]
    \For{each \textit{sample} in \textit{dataset}}
        \State Compute \textit{max\_output\_size}, \textit{output\_mean}, \textit{output\_std}
        \State Determine \textit{has\_max\_length\_degeneration}
        \State Determine \textit{has\_high\_variance\_degeneration}
        \State \textit{has\_degeneration} $\gets$ \textit{has\_max\_length\_degeneration} or \textit{has\_high\_variance\_degeneration}
        \If{\textit{has\_degeneration}}
            \State Add \textit{sample} to \textit{degeneration\_dataset}
            \If{sample cannot be sanitized}
                \State \textbf{continue}
            \EndIf
            \State Create \textit{sanitized\_sample} from \textit{sample}
            \State Remove degenerated outputs based on condition
            \If{sanitized outputs are non-empty}
                \State Recompute statistics (mean, std, percentiles)
                \State Mark longest response as ``TEXT DEGENERATION''
                \State Add \textit{sanitized\_sample} to \textit{sanitized\_dataset}
            \EndIf
        \Else
            \State Add original \textit{sample} to \textit{sanitized\_dataset}
        \EndIf
    \EndFor
    \State \Return \textit{degeneration\_dataset}, \textit{sanitized\_dataset}
\EndProcedure
\end{algorithmic}
\end{algorithm}

\begin{figure}
  \centering
  \begin{subfigure}[b]{0.98\linewidth}
    \includegraphics[width=\linewidth]{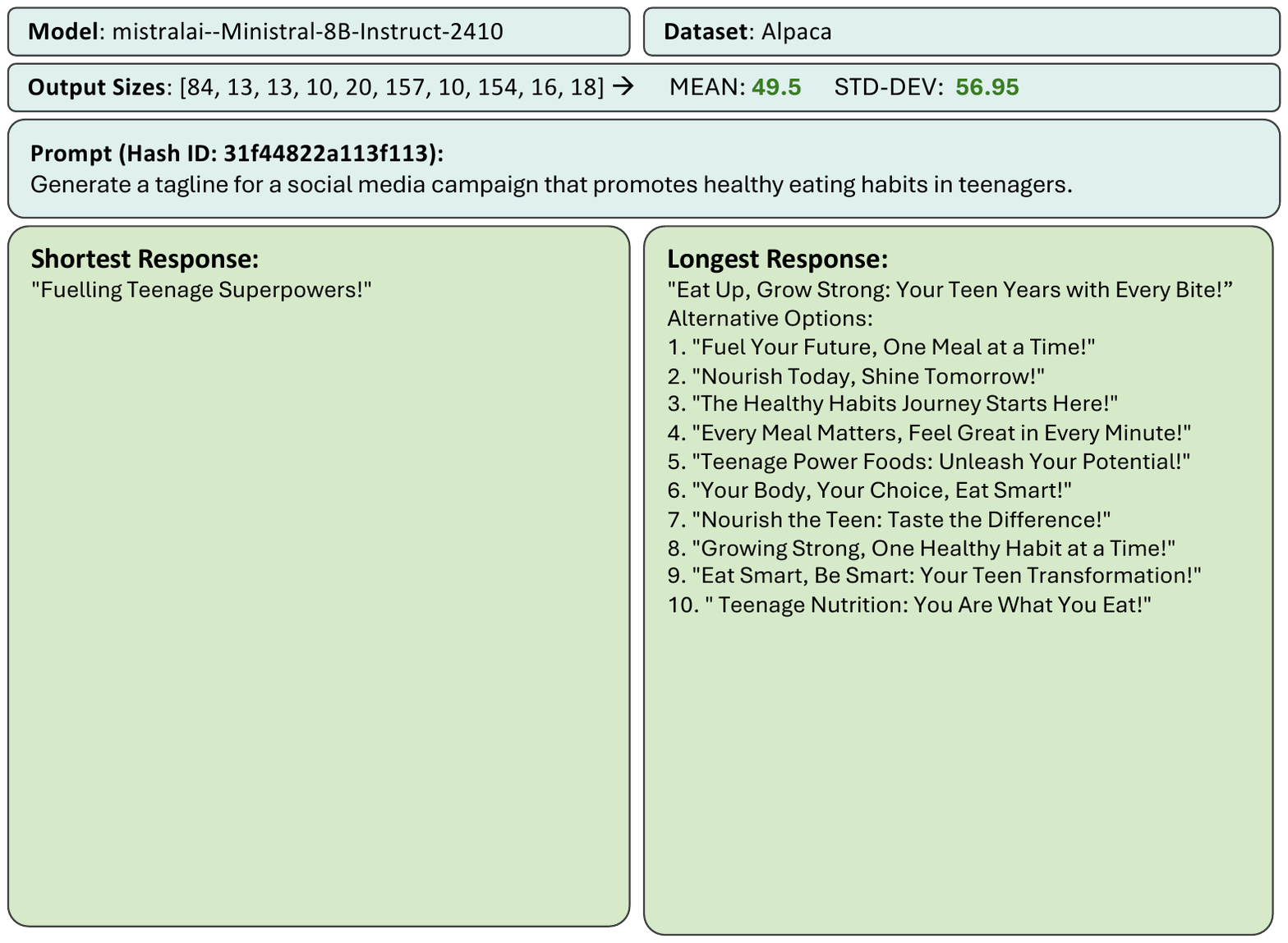}
    \caption{Difference in verbose and not-verbose output from \textbf{minist-8B} model on a sample from the \textbf{Alpaca} dataset.}
    \label{subfig:No-textdegen-2}
  \end{subfigure}
  \hfill
  \begin{subfigure}[b]{0.98\linewidth}
    \includegraphics[width=\linewidth]{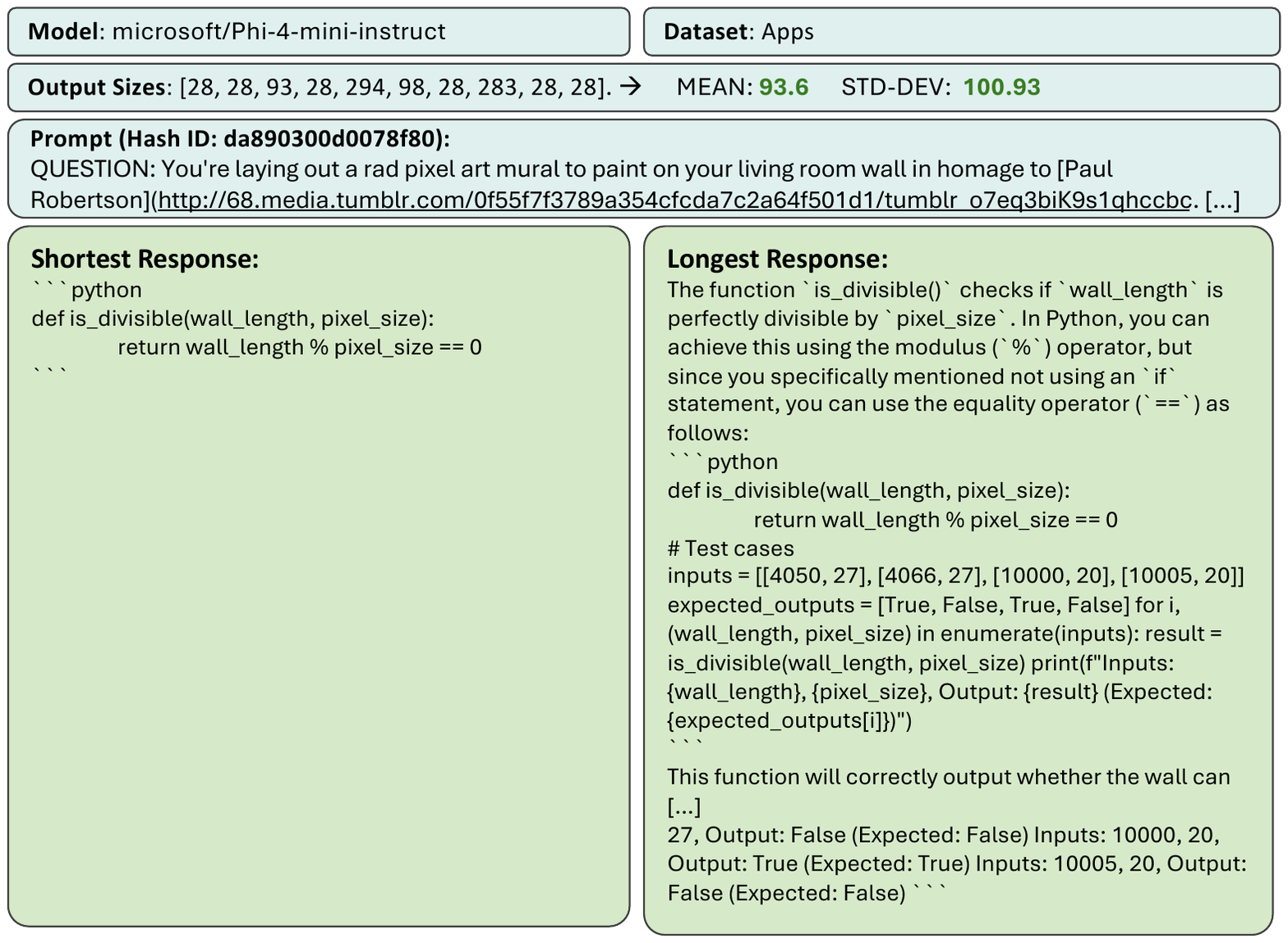}
    \caption{Difference in verbose and not-verbose output from \textbf{phi4} model on a sample from the \textbf{Apps} dataset.}
    \label{subfig:textdegen-char-letters}
  \end{subfigure}
  \caption{\textbf{NO} text degeneration examples where the standard-deviation is larger than the average in response lengths.}
  \label{appfig:no-text-degen}
\end{figure}

\begin{figure}
  \centering
  \begin{subfigure}[b]{0.97\linewidth}
    \includegraphics[width=\linewidth]{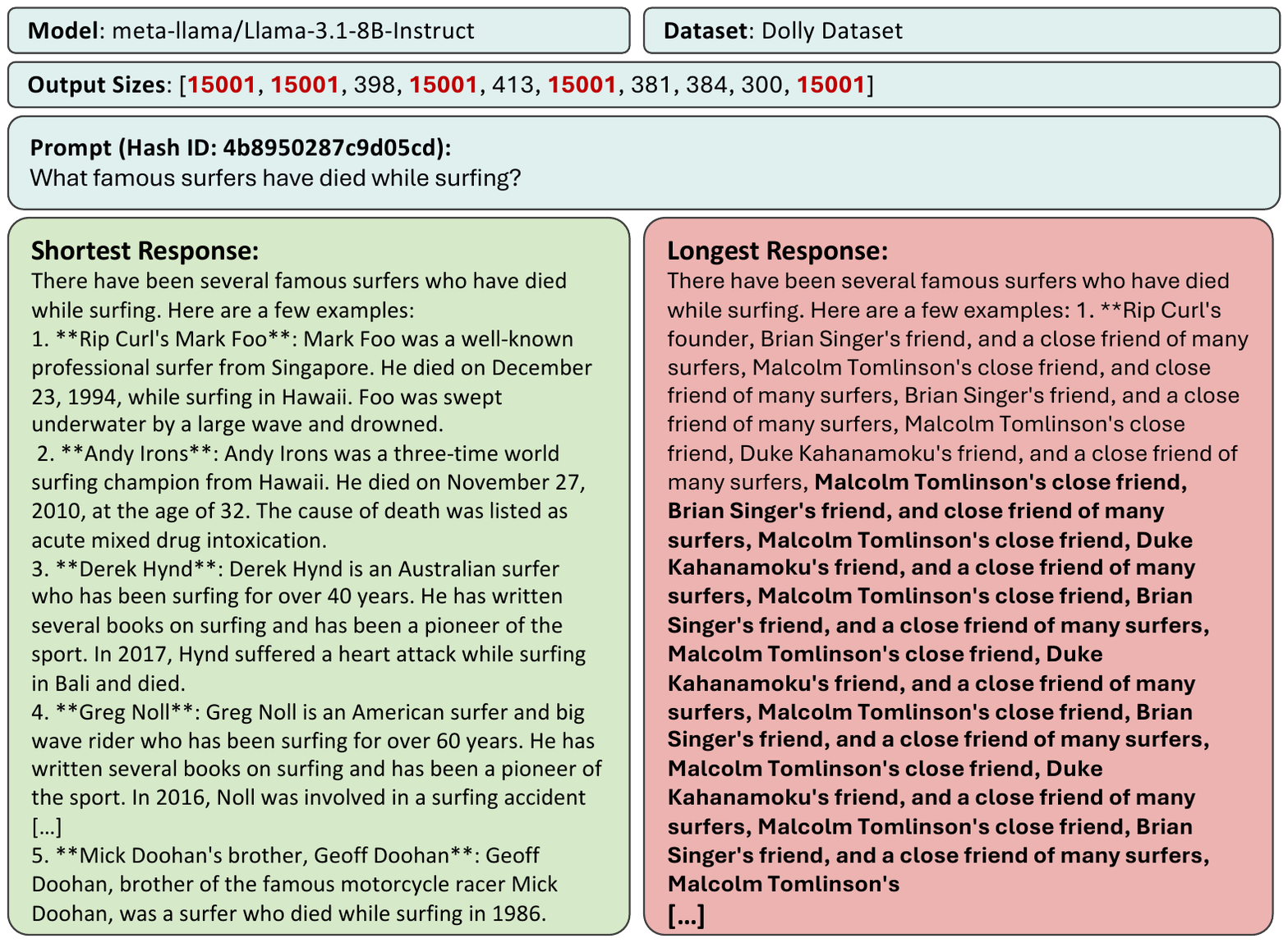}
    \caption{Sentence-level repetition of text in \textbf{llama-8B} model on a sample from the \textbf{Dolly} dataset.}
    \label{subfig:textdegen-sentence-letters}
  \end{subfigure}
  \hfill
  \begin{subfigure}[b]{0.97\linewidth}
    \includegraphics[width=\linewidth]{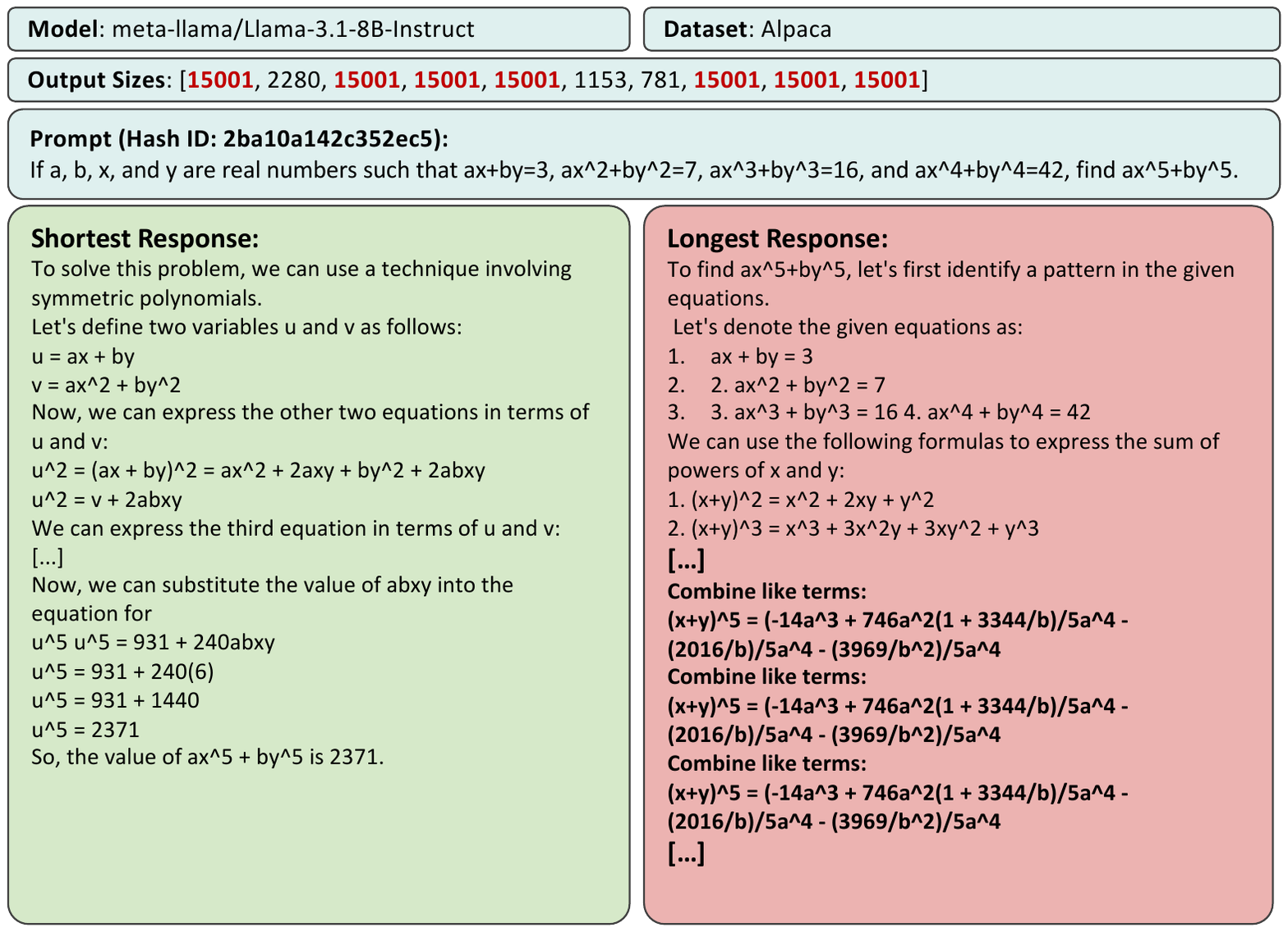}
    \caption{Block-level repetition of alphanumeric values from \textbf{llama-8B} model on a sample from the \textbf{Alpaca} dataset.}
    \label{subfig:textdegen-sentence-alphanum}
  \end{subfigure}
  \caption{\textbf{Text degeneration} examples for \textbf{sentence-level} alphanumeric text blocks.}
  \label{appfig:text-degen-sentence-block}
\end{figure}

\begin{figure}
  \centering
  \begin{subfigure}[b]{0.98\linewidth}
    \includegraphics[width=\linewidth]{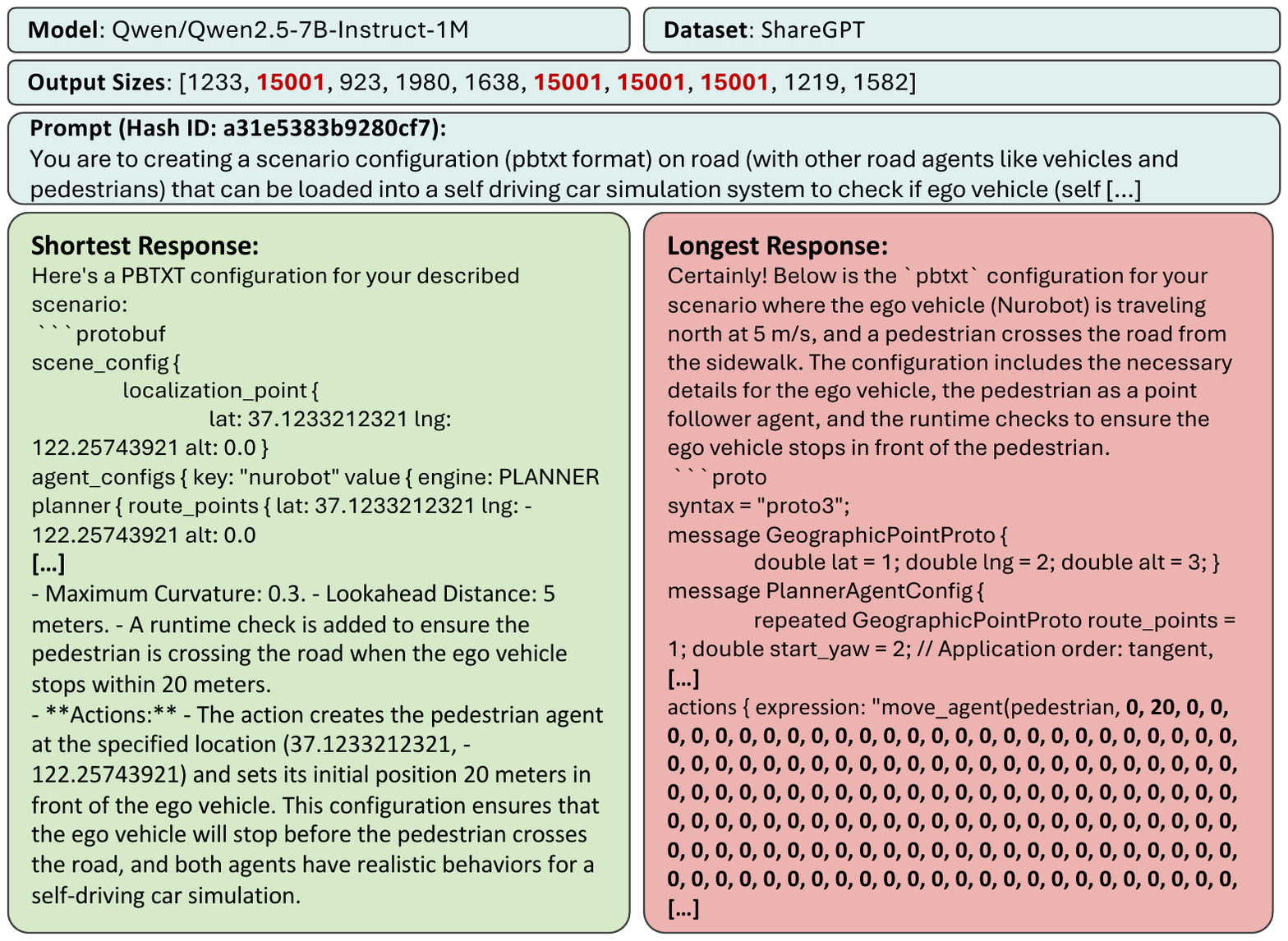}
    \caption{Character-level repetition of numbers from \textbf{Qwen-7B} model on a sample from the \textbf{ShareGPT} dataset.}
    \label{subfig:textdegen-char-numbers}
  \end{subfigure}
  \hfill
  \begin{subfigure}[b]{0.98\linewidth}
    \includegraphics[width=\linewidth]{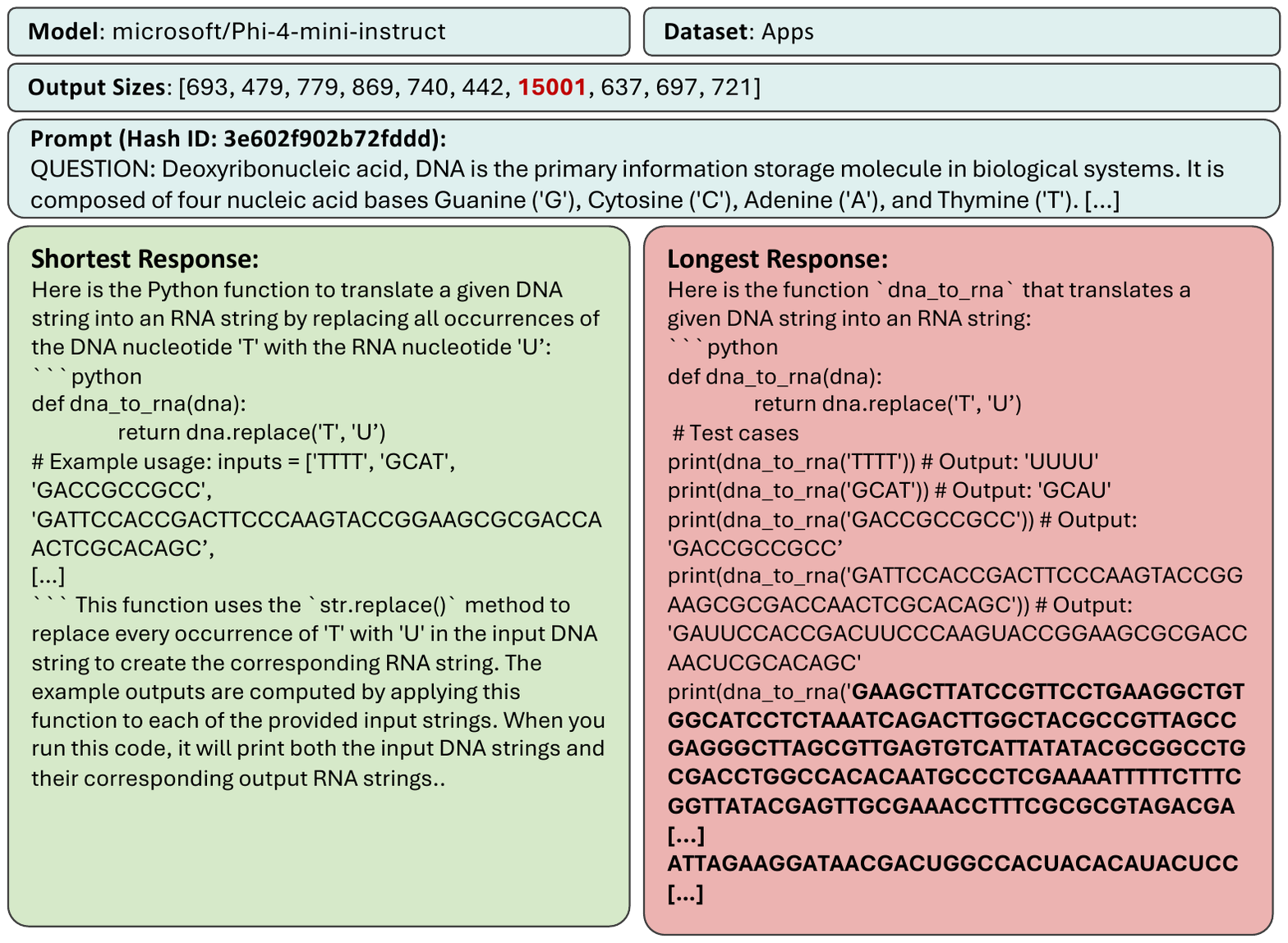}
    \caption{Character-level repetition of letters from \textbf{phi4} model on a sample from the \textbf{Apps} dataset.}
    \label{subfig:textdegen-char-letters}
  \end{subfigure}
  \caption{\textbf{Text degeneration} examples for \textbf{character-level} alphanumeric sequences.}
  \label{appfig:text-degen-char}
\end{figure}

\begin{figure}
    \centering
    \includegraphics[width=0.96\linewidth]{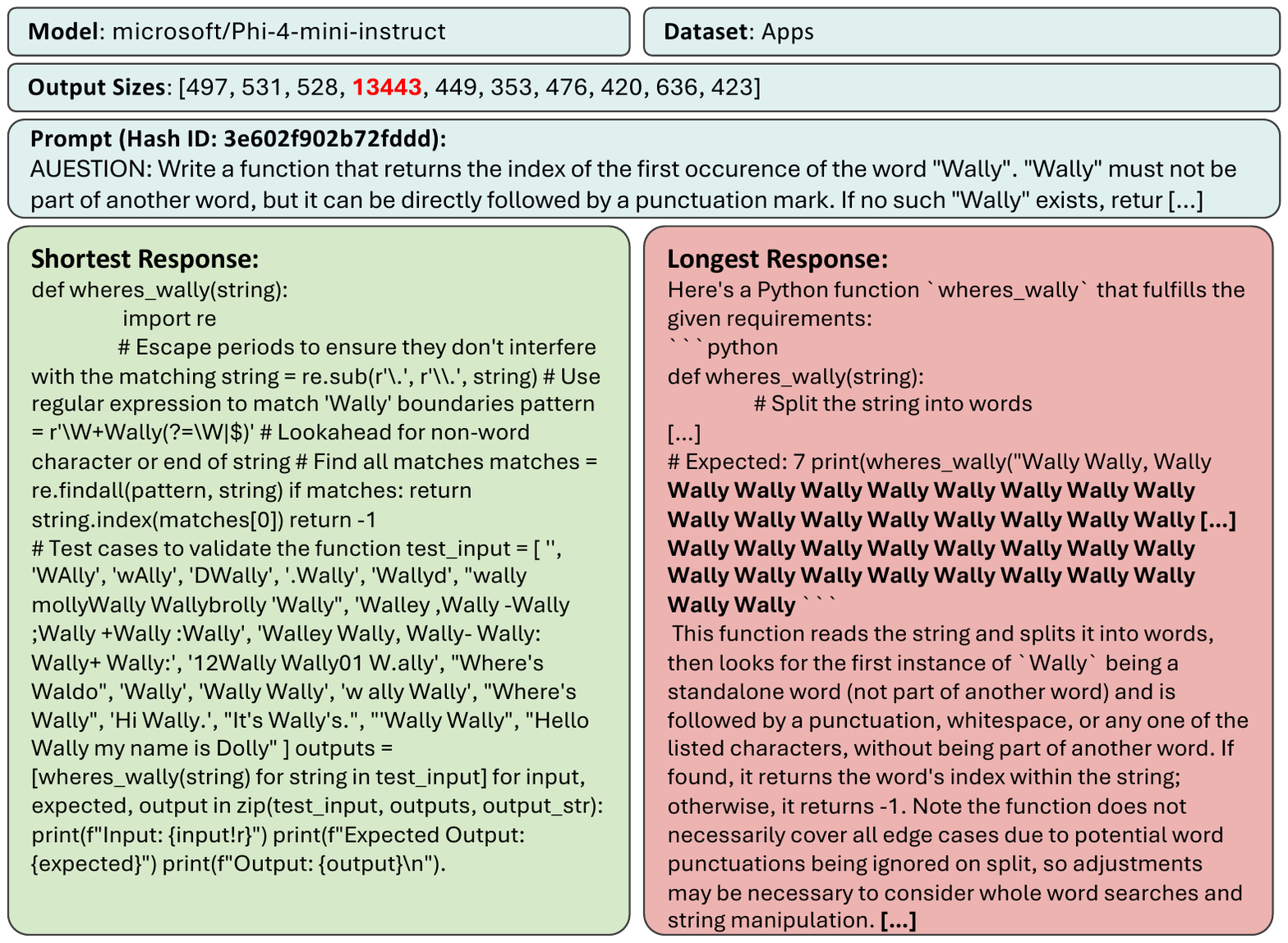}
    \caption{\textbf{Text degeneration} example of a sample that did not reach the maximum text generation output, using the model \textbf{phi-4} on a sample from the \textbf{Apps} dataset.}
    \label{appfig:text-degen-noMaxLen}
\end{figure}

This section provides further detail into how we identified and filtered instances of text degeneration in the CASTILLO default dataset, as summarized in Algorithm~\ref{alg:sanitizeddataset}.

We initially observed that certain samples exhibited unusually high standard deviations in their \texttt{output\_sizes}—in some cases, the standard deviation was greater than the mean. These cases suggested highly inconsistent response lengths across generations for the same prompt. However, upon manual inspection, we found that not all such samples were degenerated: some included valid long-form responses or variations in verbosity that did not involve repetition or other undesirable generation patterns (see Figure~\ref{appfig:no-text-degen} for an illustrative non-degenerated example).

To systematically identify text degeneration samples, we first focus on samples containing outputs that reached the maximum token limit imposed during generation (15,000 tokens). We find that all samples with one or more outputs of length $\geq$14,999 tokens contained instances of text degeneration. These instances typically exhibit either repetition at the token-level (see Figure~\ref{appfig:text-degen-char}), and sentence- or text-block level (see Figure~\ref{appfig:text-degen-sentence-block}). We thus established the first filtering heuristic: flagging any sample where at least one output in the batch breached 15,000 tokens generation.

Further empirical inspection revealed that some degenerated samples did not reach the 15,000-token cap but still contained text degeneration patterns, such as repeated numbers or characters. These were harder to detect using a simple max-length criterion. We noticed that many of these cases exhibit a high variance in output lengths and often included at least one particularly verbose response. To capture these, we introduced a second heuristic: flag any sample where \textit{both} the standard deviation of \texttt{output\_sizes} exceeded twice the mean \textit{and} where the maximum output size was at least 8,500 tokens.

Samples matching either of these two heuristics were labeled as degeneration cases and added to the degeneration dataset. For each such case, we attempted to construct a sanitized version by removing the degenerated outputs and recomputing statistics if enough clean outputs remained. When all outputs were degenerated (e.g., all reached the 15,000-token cap), the sample was excluded from the sanitized dataset.

This two-part filtering strategy allowed us to robustly detect degeneration without over-penalizing natural variation in model verbosity.

\begin{table}[t]
    \centering
    \small
    \begin{tabular}{@{}llp{8cm}@{}}
    \toprule
    \textbf{Field} & \textbf{Type} & \textbf{Description} \\
    \midrule
    \texttt{sample\_id} & string & Unique hash including prompt, model, dataset, and generation parameters \\
    \texttt{prompt\_id} & string & Hash identifying the base prompt \\
    \texttt{model} & string & LLM used for generation \\
    \texttt{dataset} & string & Name of the dataset the prompt comes from \\
    \texttt{prompt\_text} & string & Raw text of the prompt \\
    \texttt{longest\_response} & string & Longest response from the batch \\
    \texttt{shortest\_response} & string & Shortest response from the batch \\
    \texttt{input\_size} & int & Number of input tokens \\
    \texttt{output\_sizes} & List[int] & List of response lengths (token count) \\
    \texttt{output\_mean} & float & Mean of \texttt{output\_sizes} \\
    \texttt{output\_std} & float & Standard deviation of \texttt{output\_sizes} \\
    \texttt{output\_percentiles} & Dict[str, float] & 25th, 50th, 75th, 99th percentiles \\
    \texttt{top\_k} & int & Top-$k$ sampling parameter \\
    \texttt{top\_p} & float & Top-$p$ (nucleus) sampling parameter \\
    \texttt{temp} & float & Temperature used for sampling \\
    \texttt{category} & string & Prompt category (e.g., question, instruction) \\
    \texttt{gen\_time} & float & Time taken to generate the batch (seconds) \\
    \bottomrule
    \end{tabular}
    \caption{Schema of the samples generated by the CASTILLO dataset.}
    \label{tab:dataset-schema}
\end{table}

\section{CASTILLO Schema}
\label{app:schema}

Table~\ref{tab:dataset-schema} outlines the schema of the samples generated by the CASTILLO dataset. It includes metadata for each sample—such as the prompt, model, source dataset for the prompt, and generation parameters—alongside statistical features like mean, standard deviation, and percentiles of response lengths. These fields support both regression and classification tasks aimed at modeling and predicting LLM output behaviors.

\subsection{Identifying the samples in the dataset}

We use two different hashes to identify the samples of the dataset: one for identifying the prompt and another for identifying a sample within the dataset. The prompt hash is built using the string from the dataset name concatenated with the text prompt. The sample ID is generated by taking the information used by the prompt ID, and adding the model name and the list of generation configuration parameters (temperature, top-k, and top-p) in the string. 
We choose a resolution of 16 chars for saving the prompts.

\paragraph{Hash map collisions in sample IDs}

For a 64-bit hash space, the approximate number of samples before a \textbf{50\% chance of any collision} (the ``birthday bound'') is:
\[
\Tilde{n} \approx 1.2 \times \sqrt{2^{64}} \approx 5 \times 10^9
\]
We generate approximately \textbf{130,000 samples}, so to quantify the probability of any collision among \( n = 130{,}000 \) samples in a \( N=2^{64} \) space:

\[
P \approx 1 - e^{-\frac{n(n - 1)}{2 \cdot 2^{64}}} \approx 4.581 \times 10^{-10} \approx 0.00000004581\%
\]

\end{document}